\pdfoutput=1

\documentclass[11pt]{article}

\usepackage{acl}

\usepackage{times}
\usepackage{latexsym}

\usepackage[T1]{fontenc}

\usepackage[utf8]{inputenc}

\usepackage{microtype}

\usepackage{color}
\usepackage[dvipsnames]{xcolor}
\usepackage{graphicx}
\usepackage[most]{tcolorbox}
\usepackage{hyperref}
\usepackage{listings}
\usepackage{booktabs}
\usepackage{multirow}
\usepackage{array}
\usepackage{colortbl}
\usepackage{longtable}
\usepackage{amsmath}
\usepackage[toc,page]{appendix}
\usepackage{subcaption}
\usepackage{stfloats}
\usepackage{makecell}

\newcommand{\affaa}{\raisebox{-2pt}{\includegraphics[height=10pt]{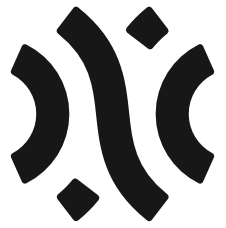}}}
\newcommand{\affhd}{\raisebox{-2pt}{\includegraphics[height=11pt]{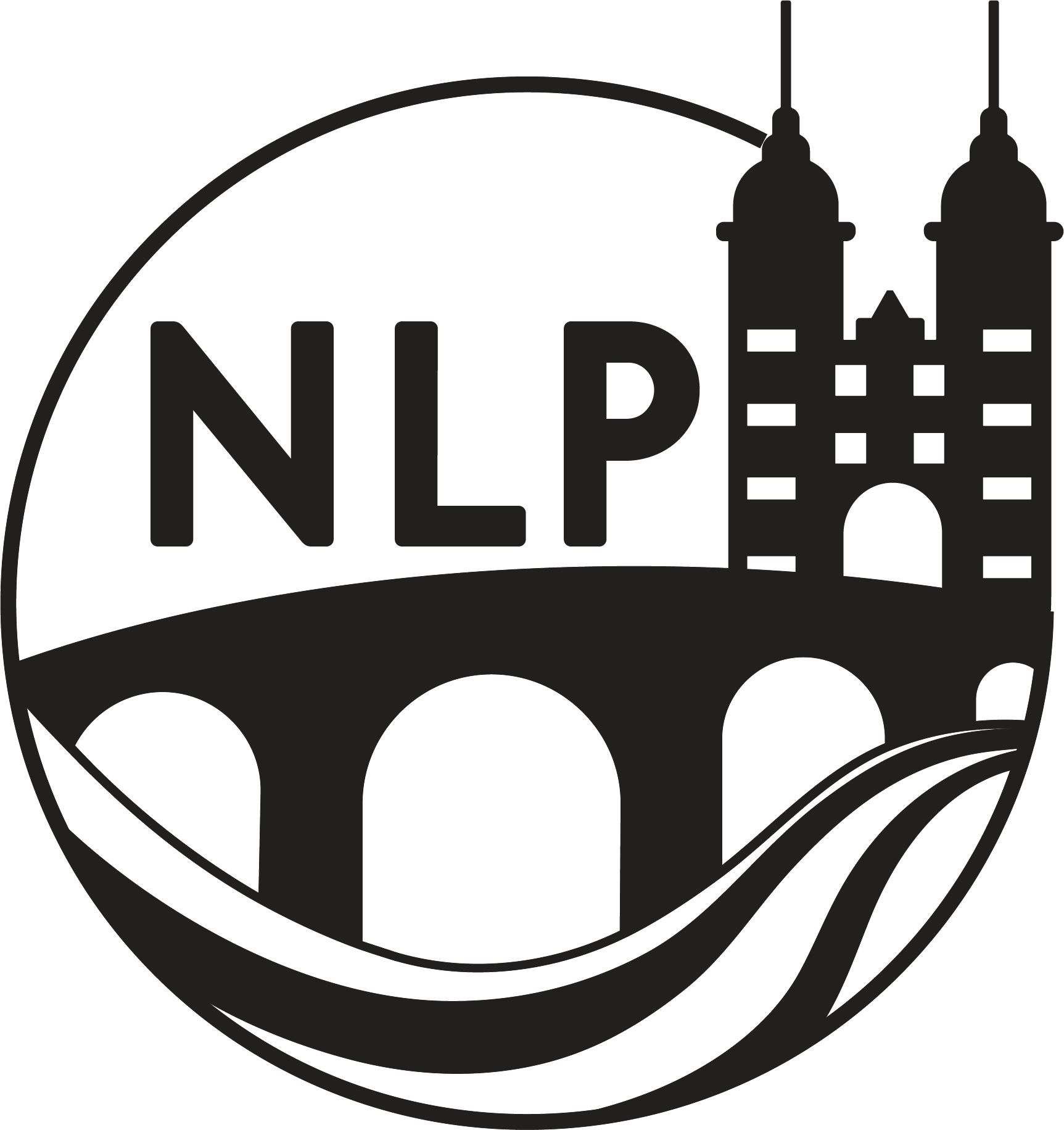}}}

\usepackage{eso-pic}
\AddToShipoutPictureFG*{%
  \AtPageUpperLeft{%
    \put(0,-25){\makebox[\paperwidth]{\small\itshape Accepted to AACL-IJCNLP 2026 (Main Conference)}}%
  }%
}

\newcommand\blfootnote[1]{%
  \begingroup
  \renewcommand\thefootnote{$\star$}%
  \stepcounter{footnote}%
  \footnotetext{#1}%
  \addtocounter{footnote}{-1}%
  \endgroup
}

\newcommand{\githublogo}{\protect\raisebox{-1pt}{\includegraphics[height=0.35cm]{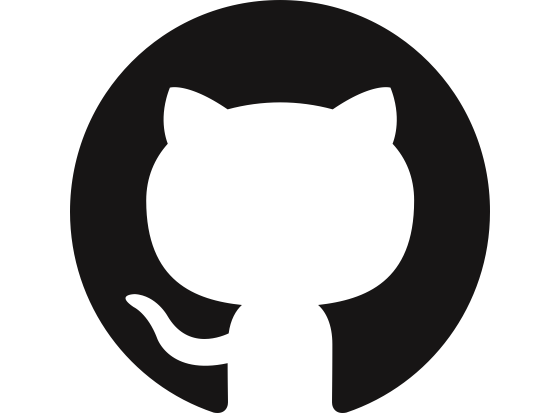}}}

\usepackage[capitalize]{cleveref}

\title{\bf GLM-RAG: Graph Language Models for Graph-Based Retrieval-Augmented Generation}

\author{
  Maya Arseven\,\affhd\,\affaa \quad
  Anette Frank\,\affhd \quad
  Beni Egressy\textsuperscript{$\star$}\,\affaa \quad
  Johann Higl\textsuperscript{$\star$}\,\affaa \quad
  Moritz Plenz\textsuperscript{$\star$}\,\affhd \\[6pt]
  \affhd\,Institute of Computational Linguistics, Heidelberg University \\
  \affaa\,Aleph Alpha Research \\[4pt]
  \small
  \texttt{\{arseven, frank, plenz\}@cl.uni-heidelberg.de} \\[-3pt]
  \small
  \texttt{\{maya.arseven, beni.egressy, johann.higl\}@aleph-alpha-research.com}
  \normalsize
}

\begin{document}

\maketitle

\blfootnote{Shared last authorship}

\begin{abstract}
Retrieval-augmented generation (RAG) over knowledge graphs requires retrievers that can effectively capture both graph structure and semantic information. Recent approaches have explored graph neural network (GNN)-based retrievers to model graph topology in multi-hop reasoning tasks. In parallel, graph language models (GLMs) have emerged as a promising paradigm that integrates graph reasoning and the semantic capabilities of language models. 
In this work, we introduce a GLM-based retriever and investigate the comparative strengths of GLM-based, GNN-based, and traditional vector-search-based retrievers in single- and multi-hop RAG settings, and with a particular focus on transferability to unseen domains. Our findings suggest that finetuned GLM retrievers generalize better out of domain, achieving SOTA on two multi-hop benchmarks. On in-domain multi-hop QA datasets they remain comparable to prior work, with promising scaling as parameters and subgraph coverage increase. GNN-based retrievers achieve higher graph coverage with an efficient training setup, whereas the vector-search baseline excels at single-hop datasets.\footnote{\href{https://github.com/Aleph-Alpha-Research/glm-rag-open}{\githublogo\hspace{0.1cm}Aleph-Alpha-Research/glm-rag-open}}
\end{abstract}

\section{Introduction}
Language models (LMs) struggle with hallucination in many downstream applications \cite{zhang} and lack domain-specific knowledge in specialized domains. 

Retrieval-augmented generation (RAG) addresses these challenges by retrieving relevant documents and providing them as additional context \cite{lewis}. Thus, domain-specific knowledge can be integrated into an LM's reasoning process, and hallucinations can be reduced by grounding the LM's answer in relevant facts. Although RAG models show strong results in many knowledge-intensive tasks, they lag behind in tasks requiring multi-hop reasoning across multiple documents \citep{multihop-rag}. 

\begin{figure}[t]
    \centering
    \includegraphics[width=0.95\linewidth]{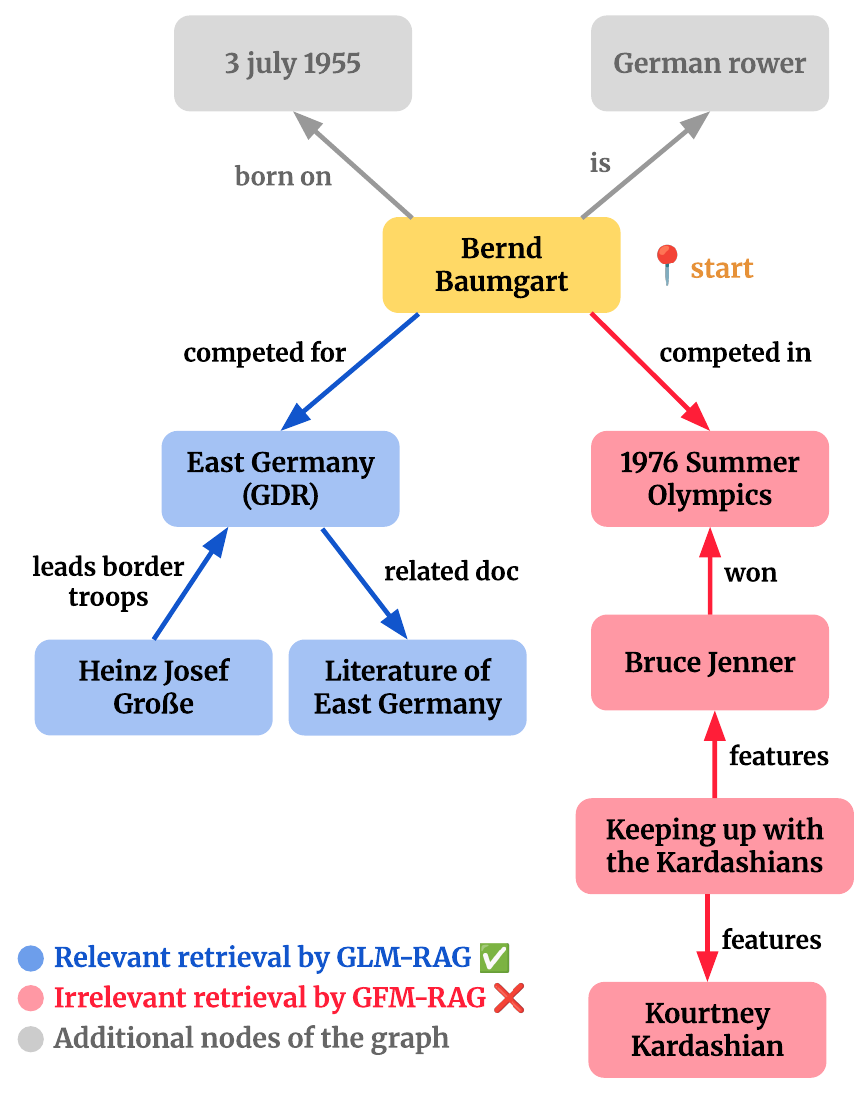}
    \caption{GLM-RAG and GFM-RAG knowledge graph retrieval traces for the query ``\textit{What is the three letter abbreviation for the country, which maintains border troops, and claims Bernd Baumgart as a citizen?}'' with ``\textit{GDR}'' being the golden answer. Example chosen from the test set of MuSiQue.}
    \label{fig:problem}
\end{figure}

Graphs can naturally capture relationships between documents. %
Unlike vector search, graph-based retrieval can traverse multiple relationships to incorporate multi-dimensional retrieval chains, enabling stronger cross-document interactions. Hence, graph RAG methods have been introduced to combine the strengths of graph structures and RAG models \cite{graphrag}.

Current graph RAG methods \citep{hipporag, gretriever, main} typically rely on shallow semantic integration, such as sentence embeddings, and are unable to support end-to-end training in a unified graph-text model. %
Figure \ref{fig:problem} shows a %
knowledge graph and %
the retrieval chain of graph RAG models for a complex query:
``\textit{What is the three letter abbreviation for the country, which maintains border troops, and claims Bernd Baumgart as a citizen?}''.

GFM-RAG (see details in §\ref{gnn-retriever}) does not utilize node labels. %
Consequently, it treats ``\textit{East Germany (GDR)}'' and ``\textit{1976 Summer Olympics}'' as equally relevant neighbors of ``\textit{Bernd Baumgart}'': both are direct neighbors of ``\textit{Bernd Baumgart}'' and connected via similar relations (``\textit{competed for}'' and ``\textit{competed in}'').
By relying on these structural signals, GFM-RAG  retrieves irrelevant entities such as ``\textit{Bruce Jenner}'' and ``\textit{Kourtney Kardashian}''.

Considering such shortcomings, we see great potential in unleashing the semantic meaning of text-attributed graphs, by using a model that has a good understanding of the meaning of text and can, at the same time, operate on graphs. Following this motivation, we propose \textbf{GLM-RAG}: \textbf{G}raph \textbf{L}anguage \textbf{M}odels for Graph-based \textbf{R}etrieval-\textbf{A}ugmented \textbf{G}eneration. Building on GFM-RAG \cite{main}, we replace their GNN and sentence-embedding-based retriever with an end-to-end trainable Graph Language Model \citep[GLM;][]{glm}. GLMs adapt pretrained language models into graph transformers while preserving their pretrained parameters, and hence language understanding capabilities (see §\ref{sec:glm}). This allows GLM-RAG to natively process text-attributed graphs on the token level, simultaneously leveraging both structural information and the semantic meaning of entities, relations, and queries.
In the above example (cf.\ also \cref{fig:problem}) , this enables GLM-RAG to make an  informed decision between the two similar neighbors and successfully retrieve the semantically relevant neighborhood.

In addition to introducing GLM-RAG, we systematically assess when graph-enhanced retrieval methods offer benefits. 
We conduct comprehensive experiments across a range of domains, enabling a rigorous comparison between graph-based methods and a vanilla vector-search approach. 

Zero-shot results on out-of-domain (OOD) single-hop datasets reveal that vanilla RAG is sufficient for datasets that require only single-step reasoning. 
This is in line with recent findings by \citet{whengraph}. However, in multi-hop settings, we see that the graph-based models outperform vanilla RAG. Furthermore, results on OOD datasets highlight GLM-RAG's strong transferability performance, outperforming GFM-RAG in all settings and achieving new SOTA results on Medical and computer science G-bench~\citep{whengraph,bench-cs}.

Beyond these advantages, our ablations reveal that GLM-RAG's performance scales well with model size, while GFM-RAG plateaus. Combined with the strong OOD performance, this shows GLM-RAG's merit as a general graph RAG foundation model.

Lastly, our analysis of the ranking results reveals that GLM-based retrievers show particular strengths on questions requiring deeper semantic understanding, whereas GNN-based retrievers profit from covering larger parts of the knowledge graph.

Our main contributions are:
\begin{enumerate}
    \item We propose \textbf{GLM-RAG}, a new graph RAG framework powered by a Graph LM-based retriever for stronger integration of text features
    \item GLM-RAG shows strong \textbf{zero-shot} generalization capabilities, surpassing comparable graph foundation models
    \item Our \textbf{comparative results} show that vanilla RAG is sufficient for single-hop retrieval tasks, while graph-based methods excel in \textit{multi-hop} settings
    \item Ablation studies show that GLM-RAG \textbf{scales better} with increasing model size compared to GNN-based retrievers
    \item Extensive \textbf{analysis} reveals that GNN-based retrievers achieve broader graph coverage but lack the stronger \textit{semantic understanding} of GLM-based retrievers

\end{enumerate}

\section{Previous Work}

\paragraph{Retrieval-Augmented Generation (RAG).} RAG typically relies on dense vector search, where text corpora are split into isolated chunks and indexed via sentence embeddings \cite{reimers-gurevych-2019-sentence}. Once retrieved, relevant documents are included in an LM's context to help answer questions \cite{lewis,karpukhin-etal-2020-dense}. 
While offering a cheaper alternative to fine-tuning models, RAG models struggle with multi-hop reasoning and complex queries that span across separate texts.

\paragraph{Graph RAG.} Graph RAG methods aim to solve more complex multi-document tasks by encoding informational dependencies within a graph. \citet{graphrag} frame this as a query-focused summarization task, rather than an explicit retrieval task. However, later works use GNN-based retrievers to leverage the graph topology of KGs in multi-hop QA tasks \cite{gnnrag}. %
Most relevant to our work is GFM-RAG \citep{main},  a ``graph foundational model'' powered by a GNN-based retriever, designed to be generalizable to unseen domains without finetuning. GFM-RAG achieved SOTA results on in-domain datasets and showed promising transferability capabilities beyond the training data. 
In recent work, \citet{greasoner} extended GFM-RAG by including more diverse information in the constructed KG, thus surpassing the previous SOTA. 

In this work, we build on GFM-RAG, with the objective of improving its transferability by using a Graph LM (GLM) for retrieval \citep{glm}. Graph LMs are based on pretrained LLMs and therefore combine a strong understanding of (unseen) texts with graph processing capabilities. Combining our approach with \citet{greasoner} is left for future work.  %

\paragraph{LMs on graphs.} A common approach to encode KGs is to linearize and process them with a LM \citep{schmitt-etal-2020-unsupervised,ribeiro-etal-2021-investigating,li-etal-2021-few,gao,yamada2026when}. While capturing the text features well, such approaches make limited use of the graph structure. Hence, GNNs in combination with often static semantic embeddings are employed \citep{lin-etal-2019-kagnet,malaviya-etal-2020-commonsense,yasunaga-2022-etal-deep,zhao-etal-2023-learning}. By slightly adjusting a LM's architecture, while maintaining its pretrained parameters, a deeper interaction between text features and graph structure can be enabled \citep{glm,egressy2025setllm,gong2026nag,vajda2026teaching}. We build on this line of work, to improve graph reasoning in graph RAG. %

\section{Preliminary: Graph Language Models}\label{sec:glm}

Graph Language Models (GLMs) bridge language models and graph transformers by converting a pretrained LM into a graph transformer %
\citep{glm}. This design enables the model to leverage pretrained language understanding from the LM while incorporating structural graph reasoning through the architectural design. 

This design makes GLMs ideal for processing text-attributed graphs, such as knowledge graphs. The method involves two main steps. First, the knowledge (sub)graph undergoes preprocessing: edge and node labels are tokenized individually and connected according to the original graph structure. %
This transformation allows each triplet to be represented as a token sequence resembling natural text, while still sharing node tokens across multiple triplets, as in a standard graph structure. %
Second, the LM's self-attention mechanism is modified to use relative positional encodings that capture distances between token pairs within the same triple, thereby enabling the model to \textit{read triplets as sequential text} -- just as a language model would. While these attention patterns, which are typical for LMs, are applied to triplets, the GLM also uses GNN-like attention patterns, which capture the overall graph structure. Together, these adaptations yield a model that is able to read a labeled graph much like contiguous language, thereby achieving strong (i) language understanding abilities and simultaneously, (ii) graph-based reasoning abilities that are well-suited for graph RAG. We refer to the original publications for more details on the method as well as experimental validation. 

\section{GLM-RAG}

Building on the ability of the GLM to jointly reason over text \textit{and} graph-structured data, we propose \textbf{GLM-RAG}, a RAG architecture that relies on a \textbf{GLM-based retriever}. 

We hypothesize that a GLM-based retriever has better retrieval quality and subsequently QA performance, due to its better understanding of text features in the graph, compared to
a GNN-based retriever, which can only utilize the graph structure with shallow text integration. 

We build on the GFM-RAG framework of \citet{main}, adopting their KG and QA dataset construction stages, as well as their document ranking and answer generation mechanisms. Our contribution is a \textbf{GLM-based retriever} (§\ref{sec:glm-retriever}) that replaces the original KG retrieval component. 
Crucially, this setup yields a controlled environment that allows us to assess a GLM-based versus a GNN-based retriever. Please refer to App. \ref{app:implementation} for more implementation details.

\subsection{KG \& QA Dataset Construction}
The pipeline starts from a document corpus, and constructs a corresponding knowledge graph on entity-level, via an LM-based named entity recognizer (NER), so that a document such as ``\textit{AACL 2026 is held in Hengqin, China.}'' is represented through the tuples \texttt{[(AACL\_2026, held\_in, Hengqin), (Hengqin, part\_of, China)]} within the KG.     

The queries undergo a similar transformation, so that a question ``\textit{Where is AACL 2026?}'' is represented through a seed entity ``\texttt{AACL\_2026}'' as a starting point within the KG. There may be multiple seed entities depending on the complexity of the query.

\subsection{KG Retrieval}\label{sec:gnnvsglm}

\begin{figure*}
    \centering
    \includegraphics[width=\linewidth]{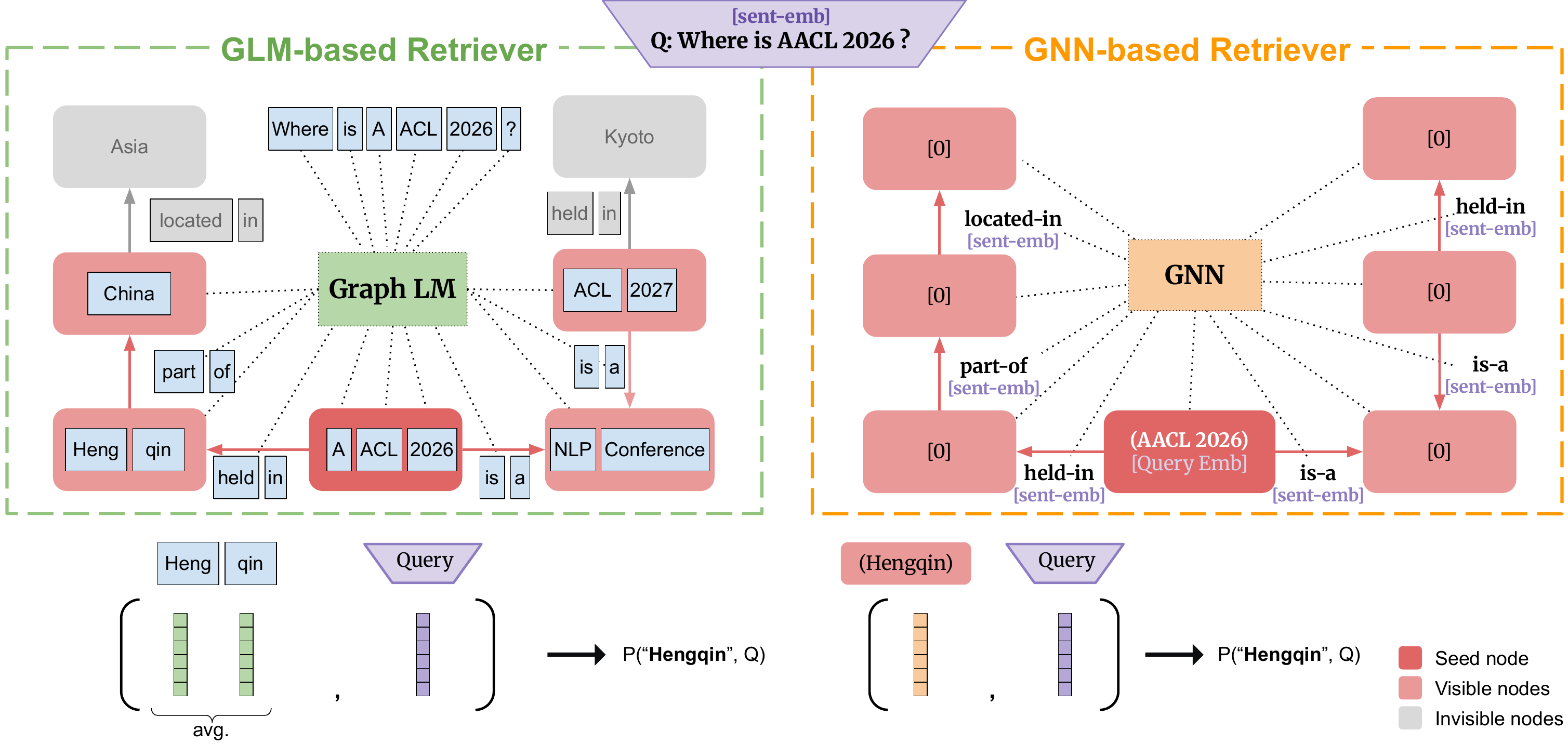}
    \caption{Differences between how GNN- and GLM-based retrievers reason over the KG for the query ``\textit{Where is AACL 2026?}''. \textcolor{Green}{GLM-based retriever (left)} tokenizes the nodes, relations, and the query with the Graph LM's encoder. The  tokenized input is passed through the graph LM to produce embeddings for each token. The token embeddings corresponding to each node are aggregated to produce the final node representation, which is used for computing the relevance scores of each entity in the extracted subgraph. 
    \textcolor{orange}{GNN-based retriever (right)} calculates the query and relation embeddings via a sentence embedding model. Only the seed entit(ies) get initialized with the query embedding and the rest of the nodes with the zero vector. The GNN iteratively propagates messages along edges using message passing over the full graph, producing a score distribution over all nodes.}
    \label{fig:glm-rag}
\end{figure*}

\subsubsection{GNN-based Retriever}\label{gnn-retriever} 

The main building block of GFM-RAG is the query-dependent GNN retriever, whose job it is to rank the entities of the constructed KG by relevance to a given input query. The message passing paradigm \cite{convgnn} is used to propagate information across the graph (see right side of Fig.~\ref{fig:glm-rag}).

While seed entities are initialized with the query embeddings, all other nodes are initialized with zero vectors. By contrast, relations are initialized with their sentence embeddings. Consequently, the resulting node representations from the GNN are driven by their relative graph distance to the seed entities rather than their semantic features. Node texts are used exclusively to identify seed entities and play no further role in the GNN retriever. Given this shortcoming, we also explore a variant of GFM-RAG, GFM-RAG+, where all nodes are initialized with their sentence embeddings (see Sec. \ref{sec:baselines}).

At the final layer, the per-node representations are concatenated with the query embedding and scored by an MLP to find the most relevant nodes. 

\citet{main} pretrain the GNN retriever on KG completion and finetune it on three Wikipedia-based multi-hop QA datasets. 

\subsubsection{GLM-based Retriever}\label{sec:glm-retriever} 

Our proposed GLM-based retriever aims for a graph RAG approach that maximizes the potential of text-attributed graphs. 

The inner workings of the GLM retriever fundamentally differ from those of the GNN-based one. Most notably, the GNN-based retriever does not make use of the rich textual features in the graph and instead relies on the query and the raw graph structure, whereas our GLM retriever fully utilizes the semantics in the graph, through the GLM's text encoding abilities (§\ref{sec:glm}).

The GLM retriever extracts a local subgraph around the seed entities, converts it into a sequence of textual triplets, and encodes the %
individually tokenized edge and node representation through a graph encoder with structure-aware relative positions (see left side of Fig.~\ref{fig:glm-rag}). Nodes and relations are tokenized and embedded using the underlying LM's own tokenizer and token embedding layer. The tokens are passed through the LM's layers, using attention masking to encode the graph structure.
This stands in contrast to the GNN retriever, where only the seed entities are initialized with non-zero embeddings. As a result, the query-graph interaction happens at a much earlier stage, namely when the question text gets fused with the extracted subgraph structure inside the GLM's first attention layer. The resulting per-entity embeddings are element-wise multiplied with a projected question vector and passed through a scoring head.

We finetune the GLM-based retriever on three Wikipedia-based multi-hop QA datasets (§\ref{sec:experiments}).

\subsection{Ranking \& Answer Generation} 
After the ranking is completed, the most relevant entities are mapped back to their original documents, thereby converting the entity ranking to a document ranking. Finally, an LM generates the answer from the query and the top-$k$ documents.

\section{Experiments}\label{sec:experiments}

In this section, we introduce our experimental setup and report the results of the following experiments:

\textbf{1.} We finetune GLM- and GNN-based retrievers on three Wikipedia-based multi-hop question-answering (QA) datasets and evaluate retrieval (§\ref{sec:retrieval}) and QA performance (§\ref{sec:qa}).

\textbf{2.} We test the finetuned GLM-based retrievers transferability on 11 OOD datasets (§\ref{sec:transfer}).

Further ablation studies (§\ref{app:ablation}) and a breakdown of the retrieval performance (§\ref{app:perhop}) per question complexity can be found in the Appendix.

\subsection{Experimental Setup}

\subsubsection*{Models}

We choose \texttt{all-mpnet-base-v2} as the sentence embedding model for GFM-RAG. Following \citet{glm}, we initialize our GLM retriever with \texttt{T5}-large model weights, which we test in different sizes (as shown in \Cref{sec:scaling}). We use \texttt{gpt-4o-mini} as the LM for all text generation.

\subsubsection*{Datasets}

We finetune the models on the train-splits of three Wikipedia-based multi-hop QA datasets:  HotPotQA \cite{hotpot}, 2WikiMultihopQA (2Wiki) \cite{2wiki}, and MuSiQue \cite{musique}. Together, these splits yield 277k question-document pairs for training\footnote{In contrast, \citet{main} use 60k Q-Doc pairs for their results. However, in their follow-up work, \citet{greasoner} publish new results trained with 277k Q-Doc pairs.}. 

To evaluate the transferability capabilities of the GLM retriever, we follow \citet{main}, who chose seven transferability datasets from various domains: TechQA \cite{techqa}, ExpertQA \cite{expertqa}, eManual \cite{emanual}, DelucionQA \cite{delucionqa} have a customer support focus, MS MARCO \cite{msmarco} and HAGRID \cite{hagrid} contain general knowledge questions, while PubmedQA \cite{pubmedqa} requires biomedical knowledge. 
However, since these datasets do not require multi-hop reasoning, they can be solved effectively using vanilla RAG. Therefore, we also include several multihop benchmarks: \textbf{Multihop-RAG} \cite{multihop-rag}, a dataset based on English news articles, G-Bench Novel and Medical \cite{whengraph} and G-Bench Computer Science (CS) \cite{bench-cs}. Please refer to App. \ref{app:dataset} for dataset statistics.

\subsubsection*{Baselines}\label{sec:baselines}

To contextualize the results of our proposed GLM-RAG method, we compare to following baselines:

\textbf{a.} A \textbf{vanilla RAG} baseline encodes the documents as well as the queries with the \texttt{all-mpnet-base-v2} embedding model, and retrieves the top-$5$ documents according to their cosine similarity. This baseline tests the need for graph RAG methods. 

\textbf{b.} As a strong comparative baseline, we use the latest release of \textbf{GFM-RAG} on \href{https://huggingface.co/rmanluo/GFM-RAG-8M}{HuggingFace}, which uses an identical graph RAG pipeline.

\textbf{c.} A \textbf{finetuned-only} version of GFM-RAG that we refer to as \textbf{GFM-RAG*}. We opt for a finetuned-only version of GFM-RAG to test the effectiveness of the pretraining stage. This is a more comparable baseline to our GLM-RAG, considering we also only perform finetuning.

\textbf{d.} A finetuned-only version of GFM-RAG where nodes are initialized with text embeddings rather than zero vectors. We refer to this as \textbf{GFM-RAG+}. Compared to GLM-RAG, this approach also utilizes node semantics, though only via static sentence embeddings.

\textbf{e.} For QA performance we additionally test a \textbf{no-context baseline} that evaluates the LM's capability to answer questions without any retrieved context.

\subsubsection*{Metrics}

We use Recall@2 at the document level to measure retrieval performance. We evaluate document-level rather than entity-level performance, since the correct retrieval of documents directly affects the downstream QA performance. For QA performance, we report exact match (EM). All metrics are established in prior work, enabling a direct comparison. We also report Recall@5 in \Cref{app:retrieval} and F1 scores in  \Cref{app:qa}. For G-Bench, we follow standard evaluation practice for each benchmark and report, evidence recall for retrieval and accuracy (answer correctness and answer score) for QA performance. We use \texttt{gpt-4o-mini} for all LLM-as-a-judge calls. Following prior work we report one run per experiment due to computational constraints and provide significance tests in \Cref{app:significancy}.

\subsection{Retrieval Performance}\label{sec:retrieval}

\begin{figure}[ht]
    \centering
    \includegraphics[width=\linewidth]{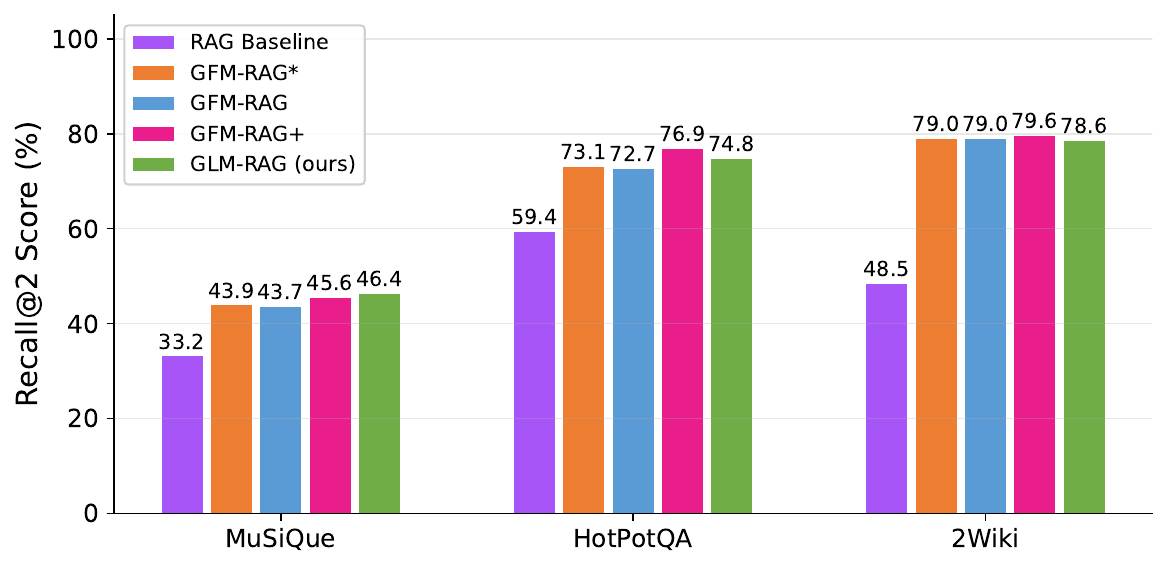}
    \caption{Retrieval performance (Recall@2) on in-domain Wikipedia datasets.}
    \label{fig:retrieval282}
\end{figure}

Figure \ref{fig:retrieval282} shows retrieval performance on the test sets of the three Wikipedia-based datasets. All graph-enhanced methods outperform the RAG baseline, highlighting the contribution of graph RAG methods in multi-hop tasks. GFM-RAG and GFM-RAG* (i.e., with and without pretraining) show near-identical performance on most datasets, calling into question the need for costly pretraining. Hence, we also only finetune our GLM retriever. 

Overall, GFM-RAG+ and GLM-RAG perform the best, showing the advantage of utilizing node semantics. On HotPotQA and 2Wiki, GFM-RAG+ slightly outperforms GLM-RAG, while GLM-RAG achieves the best result on MuSiQue. This confirms GLM-RAG's competitive in-domain performance, despite being restricted to smaller subgraphs.

\cref{tab:retrieval} shows the performance of additional graph RAG methods, including new SOTA results by \citet{greasoner} in concurrent work. Their main innovation is to extend GFM-RAG by including more information in the KG-indexing step. This is orthogonal to our work and is compatible with our proposed GLM-Retriever. Combining both approaches is beyond the scope of this paper and is left for future work.

\subsection{QA Performance}\label{sec:qa}

\begin{figure}[h]
    \centering
    \includegraphics[width=\linewidth]{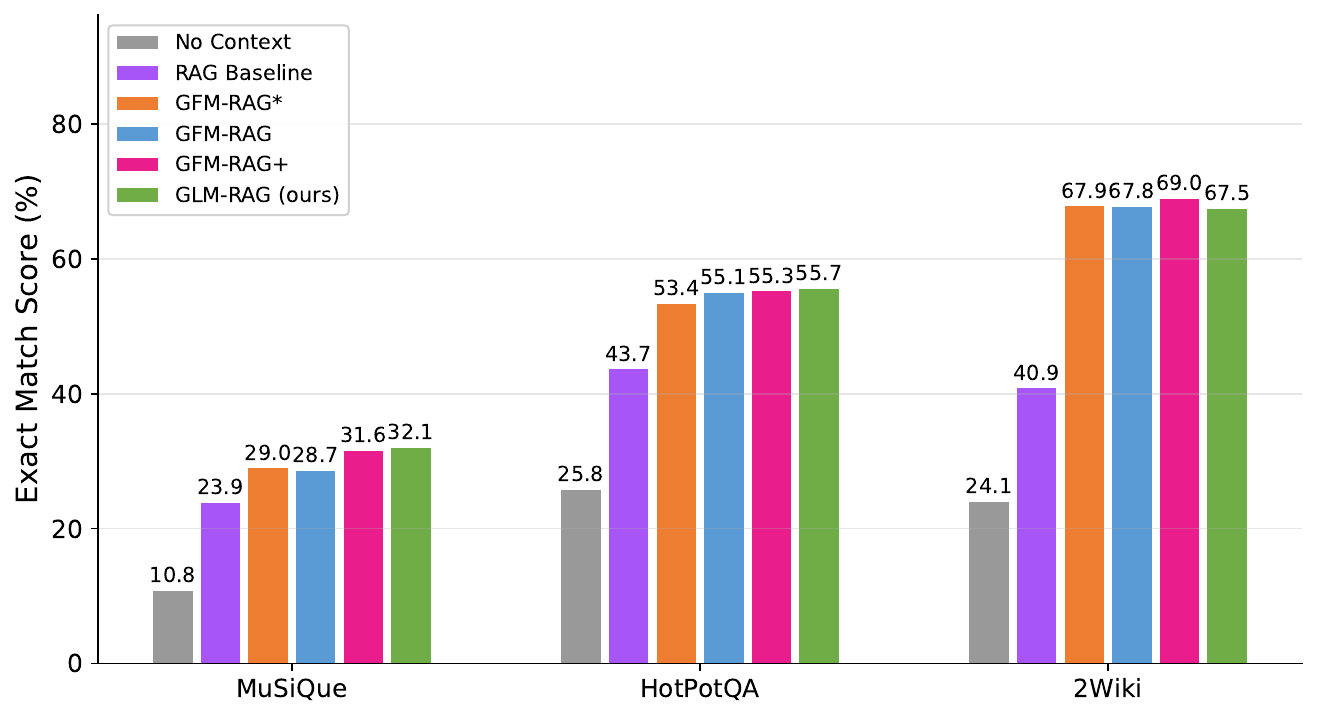}
    \caption{QA performance (exact match) on in-domain Wikipedia datasets.}
    \label{fig:qa}
\end{figure}

A similar trend can be observed in downstream QA performance in Figure \ref{fig:qa}. The no-context and RAG baselines get the lowest scores, showing the value of the retrieved context as well as graph-enhanced retrievers. GFM-RAG and GFM-RAG* perform comparably across all three datasets, with a maximum difference of 1.7 points on HotPotQA, confirming that the effect of pretraining is negligible.

The overall performance trends of GLM-RAG and the baselines are comparable to the results in retrieval. The only difference is that GLM-RAG achieves the best EM scores on HotPotQA. QA results of further methods are shown in \cref{tab:qa_results}.

\subsection{Transferability}\label{sec:transfer}

\begin{figure*}[ht]
    \centering
    \includegraphics[width=\linewidth]{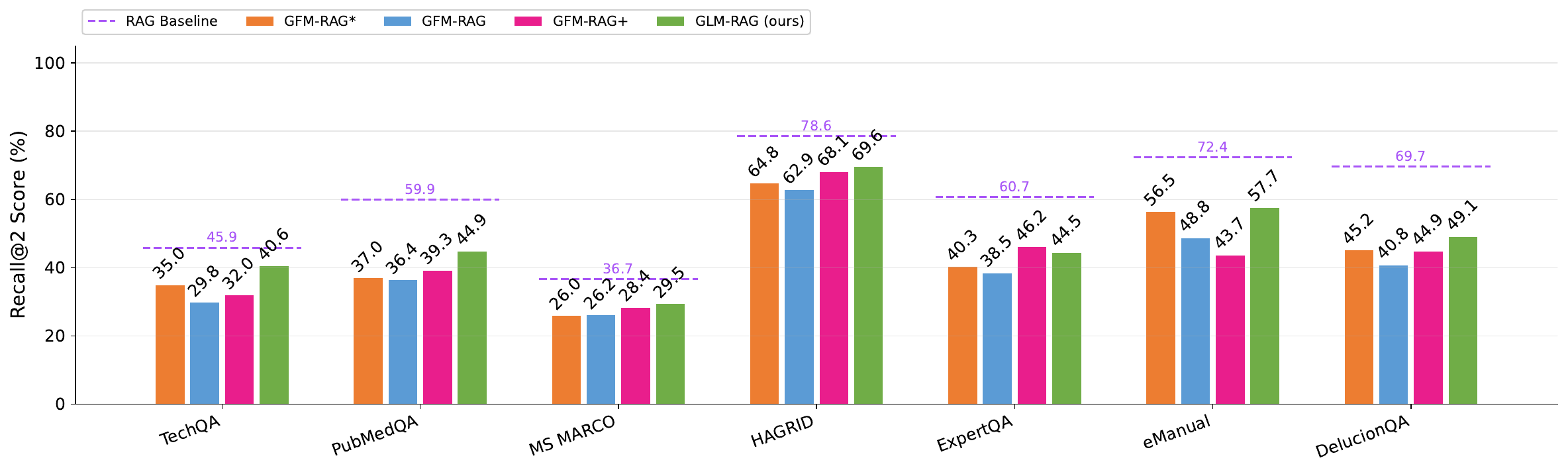}
    \caption{Retrieval performance (Recall@2) on single-hop out-of-domain datasets.}
    \label{fig:transfer}
\end{figure*}

While GLM-RAG's in-domain performance is not significantly better than GFM-RAG's, transferability experiments reveal a significantly better generalization to OOD datasets. 

In the following, we first train our models on only one of the Wikipedia datasets, and examine its transfer to the remaining Wikipedia datasets (§\ref{sec:wiki-ablation}). Then we test the fully-trained models on seven single-hop OOD datasets (§\ref{sec:transfer}), and a multi-hop OOD dataset (§\ref{sec:multihop}).

\begin{table}[ht]
    \centering
    \caption{Generalization abilities of GLM and GFM-RAG(*,+) when finetuned on one Wikipedia dataset.}
    \label{tab:wiki_generalization}
    \scriptsize
    \resizebox{\linewidth}{!}{
    \begin{tabular}{ll|ccc}
        \toprule
         &  & \multicolumn{3}{c}{ \textbf{Recall@2} } \\ \midrule
        \textbf{Train Set} & \textbf{Test Set} & \textbf{GLM-RAG} & \textbf{GFM-RAG*} & \textbf{GFM-RAG+} \\       
        \midrule
        \multirow{3}{*}{MuSiQue}  & HotpotQA             & \textbf{65.7} & 59.4 & 60.3 \\
                                  & MuSiQue  & \textbf{44.6} & 40.8 & 41.7 \\
                                  & 2Wiki                & \textbf{74.0} & 71.3 & 72.6 \\
        \midrule
        \multirow{3}{*}{HotpotQA} & HotpotQA & 74.2 & 73.5 & \textbf{77.0} \\
                                  & MuSiQue              & \textbf{45.4} & 42.6 & 43.6 \\
                                  & 2Wiki                & \textbf{76.4} & 75.6 & 75.7 \\
        \midrule
        \multirow{3}{*}{2Wiki}    & HotpotQA             & \textbf{64.3} & 57.7 & 57.8 \\
                                  & MuSiQue              & \textbf{43.3} & 39.8 & 39.1 \\
                                  & 2Wiki    & 77.7          & 78.9 & \textbf{79.6} \\
        \bottomrule
    \end{tabular}
    }
\end{table}

\subsubsection{Transferability on Wikipedia Datasets}\label{sec:wiki-ablation}

We train GLM-RAG, GFM-RAG* and GFM-RAG+ on each Wikipedia dataset and investigate transferability to the other two Wikipedia datasets. 

\cref{tab:wiki_generalization} shows that in all cases, GLM performs better in \textbf{out-of-domain} settings compared to the GFM-RAG variants, even surpassing GFM-RAG+ (i.e. GFM-RAG with additional node embeddings). For example, when trained on HotpotQA, GLM outperforms GFM-RAG* by 2.8 points and GFM-RAG+ by 1.8 points on MuSiQue. 

On MuSiQue, GLM-RAG is consistently better than both models, whereas GFM-RAG variants perform better \textbf{in-domain} on 2Wiki and HotpotQA.

\subsubsection{Transferability on Single-hop Datasets}

Motivated by previous results, we test the models' generalization abilities to unseen domains on 7 single-hop datasets. 

\Cref{fig:transfer} shows that GLM-RAG consistently outperforms GFM-RAG models, except for Recall@2 on ExpertQA. Considering that GFM-RAG is a foundation model trained to generalize well, this demonstrates the strong transferability capabilities of GLM-RAG. We hypothesize that this is largely enabled through GLM's deep integration of the KG's rich semantics. 

Still, it is important to note that a vanilla \textbf{RAG baseline} outperforms all tested graph RAG models in all datasets. While this may seem surprising, it is due to the nature of the datasets:
Single-hop datasets host direct questions that do not require multi-hop reasoning capabilities, so the RAG's ``simpler'' similarity metric is often sufficient to find relevant documents. Graph RAG methods, by contrast, show their strengths in multi-hop datasets, where retrieving documents that are similar to the question is not sufficient to answer compositional questions. 

\subsubsection{Transferability on Multi-hop Datasets}\label{sec:multihop}

We challenge the models by testing their retrieval and QA performance for zero-shot transferability on MultihopRAG \cite{multihop-rag}, G-Bench Novel, G-Bench Medical \cite{whengraph}, and G-Bench CS \cite{bench-cs}, all requiring multi-hop reasoning. Here, we also include the recently introduced SOTA method, G-Reasoner \cite{greasoner}.

\begin{table*}[ht]
\centering
\caption{Retrieval (Recall@2 and evidence recall) and QA (accuracy) performance on multi-hop out-of-domain datasets.}
\label{tab:multihop}
\setlength{\tabcolsep}{4pt}
\begin{tabular}{lccccccc}
\toprule
 & \multicolumn{1}{c}{\textbf{MultihopRAG}} & \multicolumn{2}{c}{\textbf{G-Bench (Novel)}} & \multicolumn{2}{c}{\textbf{G-Bench (Medical)}} & \multicolumn{1}{c}{\textbf{G-Bench (CS)}} \\
\cmidrule(lr){2-2} \cmidrule(lr){3-4} \cmidrule(lr){5-6} \cmidrule(lr){7-7}
\textbf{Method} & R@2 & Recall & ACC & Recall & ACC & ACC \\
\midrule
RAG        & 32.5 & 55.9 & 47.9 & 75.1 & 61.0 & 71.7 \\
GFM-RAG    & 34.1 & 75.9 & 58.6  & 82.2 & 72.2 & 72.1 \\
GFM-RAG+   & 39.0 & 86.1 & 60.6 & 93.3 & 75.7 & 76.1 \\
G-Reasoner & 34.9 & 87.7 & 58.9 & 93.8 & 73.3 & 73.9 \\
GLM-RAG    & \textbf{60.0} & \textbf{88.0} & \textbf{61.9} & \textbf{94.6} & \textbf{76.9} & \textbf{76.6} \\
\bottomrule
\end{tabular}

\end{table*}

\Cref{tab:multihop} shows the retrieval and QA results for all methods. GLM-RAG's retrieval performance excels in this multi-hop zero-shot setting, surpassing the RAG baseline, GFM-RAG variants and G-Reasoner by 20 points in Recall@2 on MultihopRAG. The same holds for the three G-Bench benchmarks, where GLM-RAG outperforms all baselines including G-Reasoner. As of now, GLM-RAG achieves SOTA results for the Medical\footnote{\url{https://graphrag-bench.github.io/}\label{fn:shared}}  and CS\footnote{\url{https://deep-polyu.github.io/RAG/}} benchmarks, and ranks second on the Novel\footref{fn:shared} benchmark. AutoPrunedRetriever \cite{wang2026pruning}, a graph RAG method that persists the minimal reasoning subgraph built for earlier questions and incrementally extends it for later ones, is the current SOTA method for the Novel benchmark. Integrating this pruning idea to GLM-RAG is an extension that we leave out for future work.

This challenging test case underlines GLM-RAG's merit as a graph foundation model with strong capabilities for zero-shot multi-hop settings.

To conclude the main experiments, our findings suggest that even though GLM-RAG outperforms other graph-enhanced methods on the transfer datasets, a vanilla RAG is sufficient in single-step reasoning. However, on a multi-hop OOD dataset, GLM-RAG clearly outperforms both the vanilla RAG baseline and all GFM-RAG variants.

\subsection{Ablations}\label{sec:scaling}

We aim to show that GLM-RAG’s transferability gains stem from its deeper integration of textual features rather than input filtering or model scale. To test this, the following section ablates the two primary differences between GLM-RAG and GFM-RAG variants: graph input and model size. For these ablations we train all models on a smaller subset of the training data with 60k instances. %

\subsection*{Effect of Subgraph Restriction}

One of the key differences between GLM-RAG and GFM-RAG is the restricted graph visibility of GLM-RAG due to context size limitations. Here we test whether this restriction acts as a information bottleneck or a noise filter for retrieval performance. We evaluate this by adjusting the number of visible triplets in the 2-hop neighborhood subgraph in GLM-RAG as well as GFM-RAG*.

\Cref{tab:restricted_gnn} shows that GFM-RAG*'s performance is negatively affected by this restriction. Moreover, as we tighten the restriction for both models, the recall drops monotonically. This supports our claim that GLM-RAG's restricted visibility \textbf{limits} rather than aids its performance, and that relaxing it would improve results.

\begin{table}[ht]
\centering
\caption{Retrieval results (Avg. Recall@2) with restricted GFM-RAG* and GLM-RAG. Recall results are averaged over HotpotQA, MuSiQue and 2Wiki test sets.}
\label{tab:restricted_gnn}
\small
\begin{tabular}{lcc}
\toprule
\setlength{\tabcolsep}{4pt}
& \textbf{GFM-RAG*} & \textbf{GLM-RAG} \\ \midrule
unrestricted & \textbf{63.50} & - \\
600 triplets & 60.36 & \textbf{63.3} \\
500 triplets & 59.52 & 62.8 \\
400 triplets & 59.46 & 61.8 \\
300 triplets & 58.61 & 60.8 \\
\bottomrule
\end{tabular}
\end{table}

\subsection*{Effect of Model Size}

One might also argue that the gains of GLM-RAG is due to its higher parameter count, considering that the GFM* and GLM retrievers have 8M and 336M parameters, respectively. To rule out this hypothesis, we compare the models in a \textbf{capacity-matched} setting by increasing the GNN-based retrievers hidden dimensions. This results in three capacity matched pairs, shown in \Cref{tab:parameter-count}. 

\begin{figure}[ht]
    \centering
    \includegraphics[width=\linewidth]{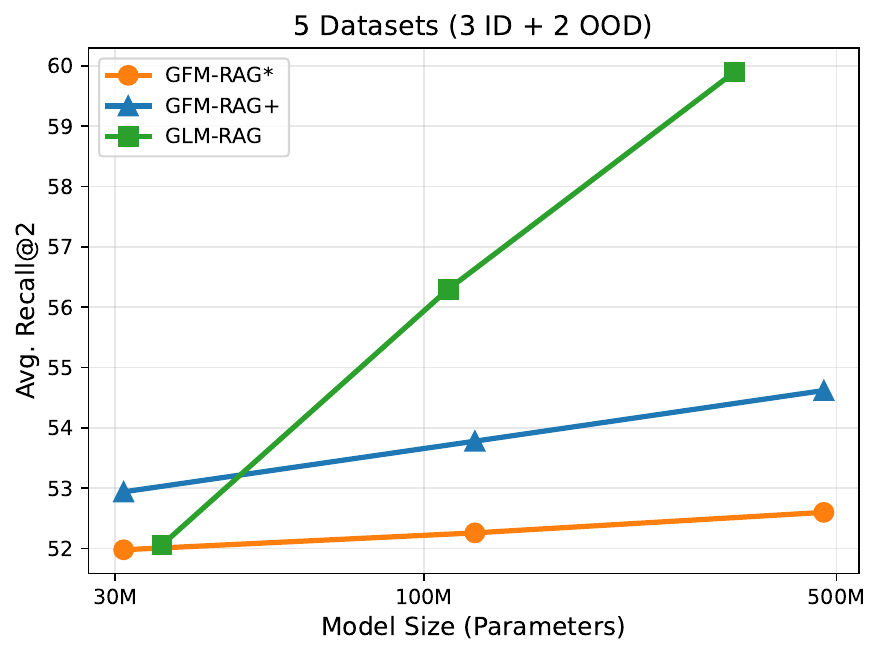}
    \caption{Model scaling comparison of GLM-RAG vs capacity-matched GFM-RAG* and GFM-RAG+. }
    \label{fig:model-size}
\end{figure}

\Cref{fig:model-size} shows the scaling trend of GNN- and GLM-based retrievers with increasing parameter count. While GFM-RAG* doesn't show any performance gains from having more parameters, GFM-RAG+'s upward trend only holds for in-domain datasets (see also \Cref{fig:gnn-vs-glm-scale} and \Cref{tab:gnn-up}). In contrast, GLM-RAG shows an upward trend in both settings, proving that its transferability advantage holds even when controlling for model size.

This finding is in line with work related to GNN's overfitting and oversmoothing problems \citep{Li_Han_Wu_2018, rong2020dropedge, oono2020graph, chen2022bag}, indicating that scaling is lacking behind transformer based models \cite{kaplan2020scaling, scaling-vt}.

To conclude, we showed that GLM-RAG's transferability claims still hold under matched visibility and capacity, implying that the generalizability gains stem from utilizing the existing large-scale pretrained LMs. %
The scalability potential of GLMs is an additional benefit that strongly indicates potential for further improvements from switching to larger and more recent LMs as the GLM's base.

\section{Analysis}\label{sec:analysis}

After evaluating the in- and out-of-domain retrieval and QA quality of the compared methods, we further investigate the strengths and weaknesses of each graph-based retriever model.
Although these properties are not a proxy for better retrieval, we analyze them to better understand the characteristics of each model's retrieval strategy. Our analysis shows that GLM uses more semantic knowledge to retrieve the right document, whereas GNN relies more on structural signals. Appendix \ref{app:ext-analysis} shows the results and calculations for these measures.

\subsection{Similarity Measures}

First of all, we want to confirm our hypothesis that the GLM-Retriever utilizes semantic features more effectively than the other models. For this analysis we compute the \textbf{semantic similarity} between the question and the retrieved answer entities. \cref{tab:semantic_sim_comparison} shows that GLM-RAG and GFM-RAG+ consistently retrieves more semantically similar entities across 11 datasets. GLM's strength is especially highlighted in the multi-hop datasets, having better semantic understanding of the constructed graphs.

\begin{table}[ht]
\centering
\caption{Comparison of cosine similarity between the question and the retrieved entities.}
\label{tab:semantic_sim_comparison}
\resizebox{\columnwidth}{!}{%
\begin{tabular}{l|c|ccc}
\toprule
\textbf{Models} & GLM-RAG & GFM-RAG*  & GFM-RAG & GFM-RAG+ \\
\midrule
\multicolumn{5}{c}{\cellcolor[HTML]{C0C0C0}\textbf{Multi-Hop Datasets}} \\ \midrule
MuSiQue & 0.309 & 0.283 & 0.286 & 0.297 \\
HotPotQA & 0.312 & 0.292 & 0.293 & 0.298 \\
2Wiki & 0.285 & 0.262 & 0.265 & 0.268 \\
MultiHopRAG & 0.245 & 0.209 & 0.212 & 0.241 \\
\midrule
\multicolumn{5}{c}{\cellcolor[HTML]{C0C0C0}\textbf{Single-Hop Datasets}} \\ \midrule
TechQA & 0.255 & 0.234 & 0.228 & 0.245 \\
PubMedQA & 0.336 & 0.298 & 0.289 & 0.314 \\
MS MARCO & 0.294 & 0.280 & 0.280 & 0.314 \\
HAGRID & 0.345 & 0.329 & 0.323 & 0.355 \\
ExpertQA & 0.284 & 0.249 & 0.235 & 0.286 \\
eManual & 0.276 & 0.234 & 0.221 & 0.286 \\
DelucionQA & 0.294 & 0.287 & 0.286 & 0.313 \\
\bottomrule
\end{tabular}
}
\end{table}

\subsection{Distance Measures}

To compare the \textbf{graph coverage} for both models, we find the shortest path from each retrieved entity to any question entity, aggregated per-sample and per-dataset.
The results in \cref{tab:mean_hop_dist} show that the GNN-based retrievers reach further on average compared to the GLM-based retriever. Although this is expected due to their architectural differences, it shows that a strength of GNN-based retrievers lies in their wider coverage of the graph.

\section{Conclusion}

In this paper, we present an alternative approach to handle multi-hop questions in graph RAG settings by training a GLM-based retriever. Our experiments show that GLM-RAG is a more generalizable method with better scalability, in line with works on building a graph foundational model. By integrating the semantic knowledge hosted in a text-attributed graph, a GLM-based retriever can better understand and select relevant documents, relying more on assessing the similarity between document entities and the question, whereas concurrent GNN-based retrievers rely more on structural signals. Our extensive comparison of vanilla RAG, GNN-based and GLM-based retrievers reveal that: (i) for single-hop questions, which do not require multi-hop reasoning, a vanilla RAG baseline is sufficient, (ii) for multi-hop questions a finetuned GFM-RAG model initialized with node embeddings (GFM-RAG+) shows competitive performance with GLM-RAG, and (iii) for zero-shot multi-hop questions, GLM-RAG shows consistently better transferability capabilities compared to its GNN-based counterparts.

\section*{Limitations} 

We build on GFM-RAG \citep{main}, which was the SOTA at the time of our experiments. Recently, \citet{greasoner} extended GFM-RAG’s KG indexing to capture more diverse information, establishing a new SOTA. Although we do not combine our GLM retriever with this indexing strategy, we expect the two approaches to be complementary, making their integration a promising direction for future research.

We have tested our approach with LMs of up to 0.8B parameters, but based on our scaling experiments, we see great potential in using much larger base LMs, but this is left for future work.

Finally, being transformer-based, our GLM-based retriever is computationally more demanding than a GNN-based retriever. Consequently, we restrict the size of the processed subgraphs to enable more efficient training. Future work could scale up GLM-based retrievers to combine their strong semantic capabilities with broader graph coverage. 

\section*{Acknowledgments}

We sincerely thank Fabien Benureau for all the contributions he made to the project.

\bibliography{anthology,custom}
\bibliographystyle{acl_natbib}

\appendix

\section*{Appendix}

\section{Details on Datasets}\label{app:dataset}

\Cref{tab:train_dataset} shows statistics of the training datasets and provides a high-level overview of the constructed KGs, including the total number of queries, documents, entities, relations, and triplets.
The latest version of GFM-RAG in \cite{greasoner} uses the full training dataset. In this work, we use a validation set for hyperparameter tuning and therefore train on 1k fewer queries per dataset. As a result, we use a total of 273,830 queries and 2,208,920 documents for training.
We also provide statistics for the test datasets in \Cref{tab:test_dataset}. The in-domain test sets each contain 1k queries, whereas the out-of-domain test sets contain varying numbers of queries, ranging from 132 (EManual) to 2,255 (MultiHopRAG).

\Cref{tab:train_graph,tab:test_graph} provide more detailed statistics about the constructed KGs using the following metrics:

The \textbf{average degree} is the mean number of edges incident to each node within the graph.

The \textbf{density} measures the ratio of actual edges to the total number of possible edges in a simple graph. Given the high values of $|V|$ in these datasets, density values near 0.0001 indicate highly sparse graphs, which is characteristic of large-scale knowledge bases where only specific and meaningful relations exist.

The \textbf{number of components} (\# Components) indicates the count of maximal subgraphs in which any two vertices are connected to each other by paths, but which are disconnected from the rest of the graph. It reflects the level of fragmentation within the dataset's knowledge structure.

The \textbf{largest connected component} (CC in \%) is the ratio of nodes in the graph's largest connected subgraph to the total number of nodes. The high percentages (approximately 99.5\%) reported across all datasets indicate that the vast majority of the knowledge space is reachable through path-based traversal.

\section{Supplementary Implementation Details}\label{app:implementation}

\Cref{tab:settings} shows the implementation and training settings of the GFM-RAG, GFM-RAG*, GFM-RAG+, and GLM-RAG models. To highlight the most important points once again:

\begin{itemize}
    \item They share the \textbf{KG-index construction} stage, since we use pre-constructed KGs for all datasets except MultiHopRAG. Only for this dataset is the \texttt{qwen-3-8b} model used for OpenIE. 
    \item The graph is represented via relative position encoding with additional buckets, depending on the setting. The default setting is \texttt{global} with \texttt{FullyConnected} enabled, resulting in three additional buckets.
    \item After hyperparameter search, the loss weights were set to 0.44 for the binary cross-entropy (BCE) loss and 0.56 for the list cross-entropy (ListCE) loss in the GLM retriever. 
    \item The GLM retriever uses differential learning rates: 5e-4 for the head (\texttt{entity\_scorer}, \texttt{question\_proj}) and 1e-4 for the T5 backbone.    
    \item The GLM retriever is trained for 2 epochs with a batch size of 2, whereas the GNN retriever is trained for 5 epochs with a batch size of 4. However, the batch size was also decreased to 2 for GFM-RAG+ due to the increased memory usage caused by including node embeddings.
\end{itemize}

\subsection{GFM-RAG+ Details}

We tested two ways to implement \textbf{GFM-RAG+} and adopt the better-performing one as a stronger baseline. Both initialize nodes with entity embeddings, however they differ in how the query is injected.

The first, ``only seed entities get query'', multiplies only the seed entities by the query embedding. The second, ``all entities get query'', initializes every node as $query\_emb * node\_emb$, contextualizing all entities with the query. This mirrors the \texttt{FullyConnected} mode of the GLM retriever, where entities also interact with the query.

In both variants we use the same embedding model for entities, queries, and relations, and the final node features combine the entity and query embeddings. GFM-RAG+ is therefore a query-dependent GNN, like GFM-RAG. As shown in \Cref{tab:gfm_rag_plus}, contextualizing all entities outperforms seeding alone on both recall metrics, so we use it as our GFM-RAG+ baseline.

\begin{table}[ht]
\centering
\caption{Different implementations of GFM-RAG+}
\label{tab:gfm_rag_plus}
\small
\begin{tabular}{lcc}
\toprule
\setlength{\tabcolsep}{4pt}
GFM-RAG+ Methods & Recall@2 & Recall@5 \\ \midrule
Only seed entities gets query & 65.7 & 79.9 \\
All entities gets query & \textbf{67.4} & \textbf{81.5} \\
\bottomrule
\end{tabular}
\end{table}

\subsection{GLM-RAG Details}

The differences between the models are largely explained in \Cref{sec:gnnvsglm}; however, we explain further implementation details for \textbf{GLM-RAG}:

\paragraph{Subgraph Selection for Scalability:} Due to its transformer-based architecture, the GLM-retriever exhibits higher computational overhead than the GNN-retriever, making it infeasible to encode an entire large-scale graph simultaneously. To address this GPU bottleneck, the GLM-retriever employs a subgraph selection strategy that scores local neighborhoods rather than the global graph. 

We control the selection ``budget'' using two hyperparameters: the number of hops (\texttt{max\_hops}) and the maximum number of triplets (\texttt{max\_triplets}). An analysis of the interaction between these parameters is provided in \Cref{app:ablation}. Based on these results, we set \texttt{max\_hops=2} and \texttt{max\_triplets=600}.

While this means that the GNN processes a larger portion of the graph, we expect that the GLM's more sophisticated ranking mechanism compensates for this restricted view.

\paragraph{Hop Prioritization:} The seed entities are extracted from the queries during the QA dataset construction process, and they serve as the starting points in the graph, similar to \citet{plenz-etal-2023-similarity}. If the \texttt{max\_hops} neighborhoods around the seed entities contain more triplets than \texttt{max\_triplets}, we sub-sample the neighborhoods to satisfy the budget constraint. We first prioritize triplets that directly connect pairs of seed entities. The remaining quota is then filled using a randomized breadth-first search starting from the seed entities. 

\paragraph{Adjustment of the Losses:} For GFM-RAG, \citet{main} employ a weighted combination of BCE and sigmoid-based ListCE (ranking) loss. Since the GLM operates only on a selected subgraph rather than the full graph, the majority of entities remain unscored with a logit of zero. In the sigmoid-based ListCE formulation \citep{listceloss}, these unscored entities still contribute to the partition function, where each zero logit contributes $\sigma(0)=0.5$. For large entity sets (e.g., $N \approx 42{,}000$), this results in a large accumulated constant contribution ($\approx 21{,}000$) in the denominator, which overwhelms the signal from positive samples and substantially weakens the learning signal. To address this issue, we mask unscored entities and restrict the normalization term only to the selected subgraph, ensuring that the model optimizes over the ranking of observed entities.

$$\mathcal{L}_{ListCE} = -\log \left( \frac{\sigma(s_{pos})}{\sum_{i \in \mathcal{V}_{s}} \sigma(s_i) + \sum_{j \in \mathcal{V}_{u}} \sigma(s_j)} \right)$$

\paragraph{Computational Costs:} GFM-RAG* is trained on 8 NVIDIA A100 GPUs (80GB) with 7 hours of supervised fine-tuning, whereas GLM-RAG \texttt{[t5-large, max\_triplets=600]} is trained on 16 GPUs with 20 hours of supervised fine-tuning. The total estimated computational cost for all experiments is approximately 180 GPU hours.

We recognize that GLM-RAG has a higher inference latency and memory usage in comparison to GNN based counterparts. \Cref{tab:inference} depicts average latency and peak GPU memory usage during retrieval in 2Wiki dataset. While optimization could potentially reduce the inference latency, we believe that the latency is acceptable for complex RAG tasks.

\begin{table}[ht]
\centering
\caption{Latency and memory usage comparison during retrieval.}
\label{tab:inference}
\small
\resizebox{\linewidth}{!}{
\begin{tabular}{lcc}
\toprule
\setlength{\tabcolsep}{4pt}
 & Latency in ms & Allocated memory in GiB \\ \midrule
GLM-RAG & 692.3 & 6.8 \\ %
GFM-RAG* & \ 18.6 & 1.3 \\ %
\bottomrule
\end{tabular}
}
\end{table}

\section{Extensive Ablations}\label{app:ablation}

\Cref{fig:scaling-recall2} shows how both the GFM-RAG* and GLM-RAG models scale with increasing amounts of training data. Following this trend, and in order to reduce computational costs, we use a smaller subset of the data consisting of approximately 60k queries and 700k documents, instead of approximately 277k queries and 2.9M documents, for the following ablation analyses:

\begin{figure}[ht]
    \centering
    \includegraphics[width=1\linewidth]{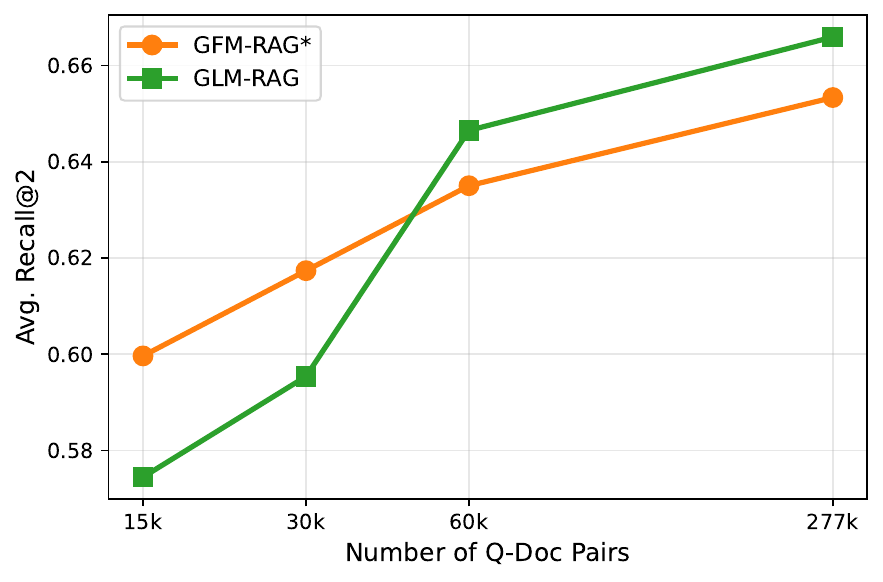}
    \caption{Data Scaling: GFM-RAG* vs GLM-RAG with increasing training data.}
    \label{fig:scaling-recall2}
\end{figure} 

In the ablations, the default GLM was initialized with \textbf{T5-base}, used \textbf{mean} embedding aggregation, and extracted a subgraph from the \textbf{2-hop} neighborhood containing up to \textbf{600 triplets}.

The query interaction was performed in two ways: by enabling the \textbf{FullyConnected} mode and by preserving the later interaction with the query embedding.

\Cref{tab:ablation} lists the ablation studies we conducted, along with the Recall@2 and Recall@5 scores averaged across the three training datasets: HotpotQA, MusiQue, and 2Wiki.

\paragraph{Embedding Aggregation:} We evaluate three methods for obtaining node embeddings from the GLM encoder's output: (i) taking the mean of the node's token embeddings, (ii) using the embedding of the first token, and (iii) using the embedding of the last token. Our experiments reveal that choosing one method over another does not lead to substantial performance differences. Therefore, we use \textbf{mean} embedding aggregation as the default setting in order to represent all tokens equally.

\paragraph{Query Interaction:} The node representations interact with the query in two places: (i) the query tokens are appended to the retrieved triplets and passed together to the T5 encoder when \texttt{use\_text} is set to \texttt{FullyConnected}, and (ii) the node representations are multiplied with the precomputed query embedding from the sentence encoder at a later stage. Disabling this later interaction is partially compensated for when \texttt{FullyConnected} remains enabled. Disabling both interactions leads to the worst results, whereas enabling both leads to the best performance; therefore, this configuration is used as the default setting.

\paragraph{Number of Hops:} The graph coverage of the GLM retriever is bounded by the number of entity-level hops it can traverse. Our experiments reveal that increasing the number of hops from 1 to 2 yields strong performance gains, whereas further increases provide only marginal improvements. Therefore, we set the retrieval scope of the model to a 2-hop neighborhood.

\paragraph{Maximum Amount of Triplets:} The extracted subgraph is also constrained by the maximum number of triplets that can be included within its 2-hop neighborhood. Here, we observe a clear trend in which retrieval performance improves as more triplets are included in the subgraph. To avoid exhausting computational resources, we ultimately set \texttt{max\_triplets} to 600; however, we hypothesize that the performance could further improve with larger triplet budgets.

\paragraph{T5-Size:} Here, we also observe a scaling trend in which retrieval performance improves with larger encoder models. Considering that the parameter sizes explored in this work are still relatively small compared to modern LLMs, which often contain tens to hundreds of billions of parameters, we hypothesize that using larger models could further benefit retrieval performance.

\paragraph{Graph-aware attention:} We implemented a baseline that linearizes the graph into a sequence of triplets and passes it through T5's encoder to rank entities. This ablates the effect of the \textbf{graph-aware attention}. Note that this discards GLM's graph-aware relative-position encoding and sparsity mask, the components that inject graph structure, leaving only T5's standard sequential positional bias. Because of linearization, the token sequence becomes longer than the graph encoding. Thus our default value of 600 triplets did not fit in memory. Consequently we trained the linearized GLM retriever with max\_triplets=300. To enable a direct comparison, we also train GLM-RAG restricted to 300 triples. In line with our expectations, the declining retrieval results show clearly that both the stronger language understanding (GLM-RAG vs GFM-RAG+), and the graph structure (GLM-RAG vs T5) contribute to GLM-RAG’s success.

\section{Extensive Retrieval Results}\label{app:retrieval}

\begin{figure*}[b]
    \centering
    \includegraphics[width=1\linewidth]{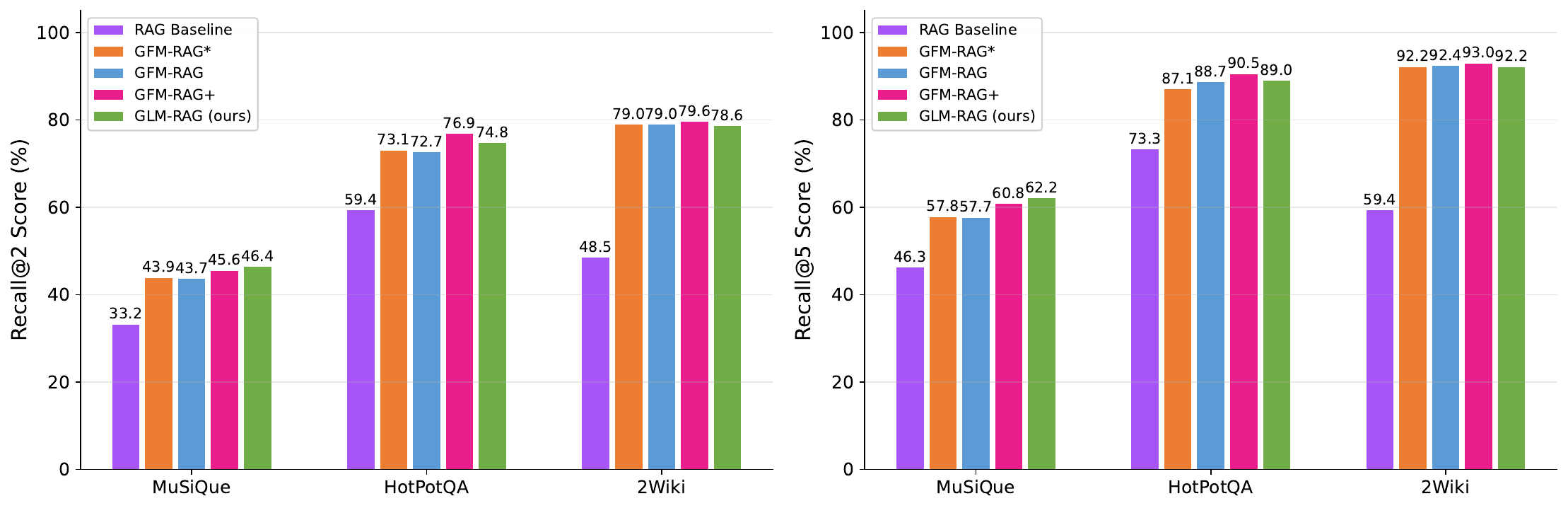}
    \caption{Retrieval performance (Recall@2 and Recall@5) on in-domain Wikipedia datasets.}
    \label{fig:ext-retrieval}
\end{figure*}

\begin{figure*}[b]
    \centering
    \includegraphics[width=1\linewidth]{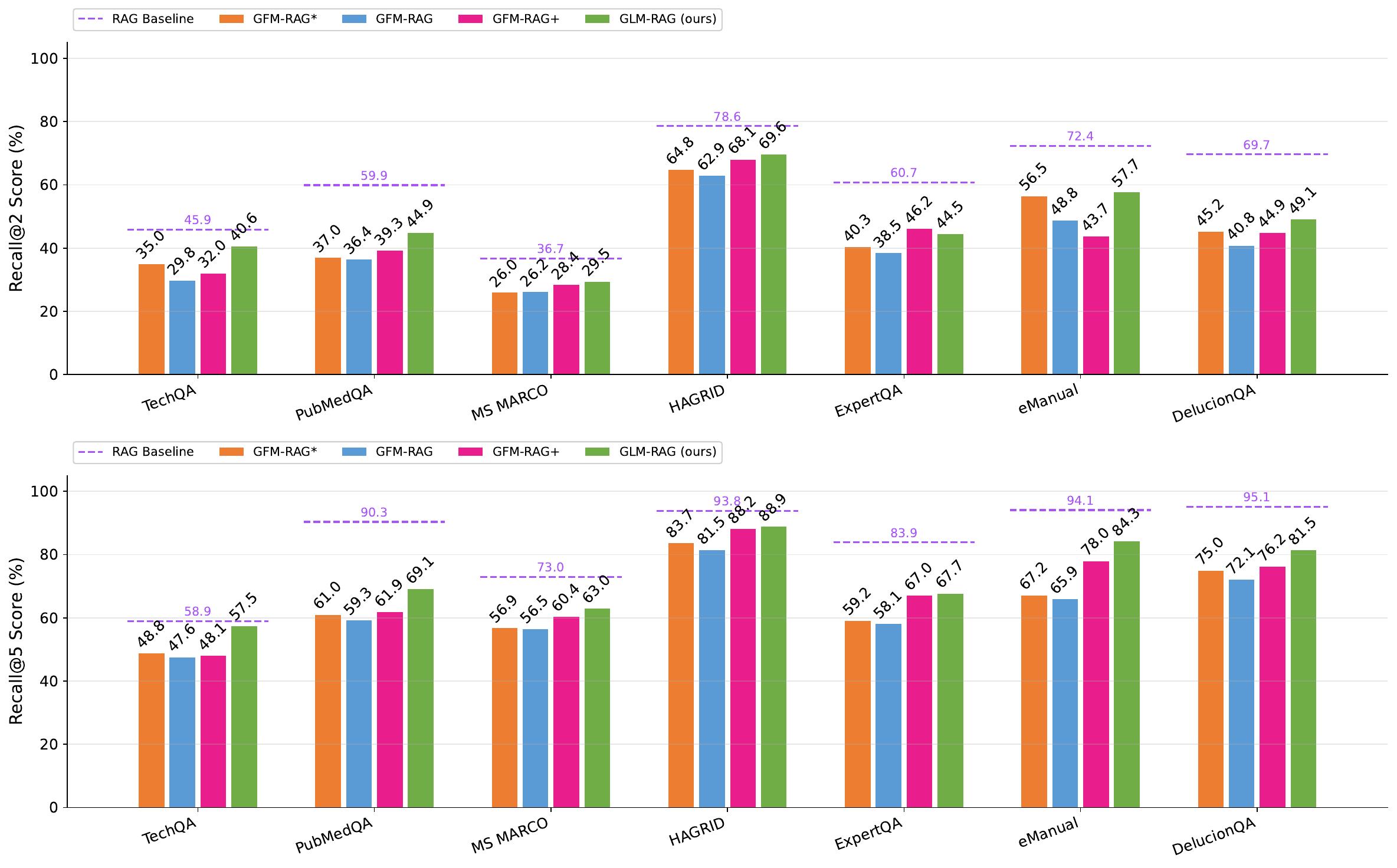}
    \caption{Retrieval performance (Recall@2 and Recall@5) on 7 out-of-domain single-hop datasets.}
    \label{fig:ext-retrieval-transfer}
\end{figure*}

\begin{figure*}[b]
    \centering
    \includegraphics[width=1\linewidth]{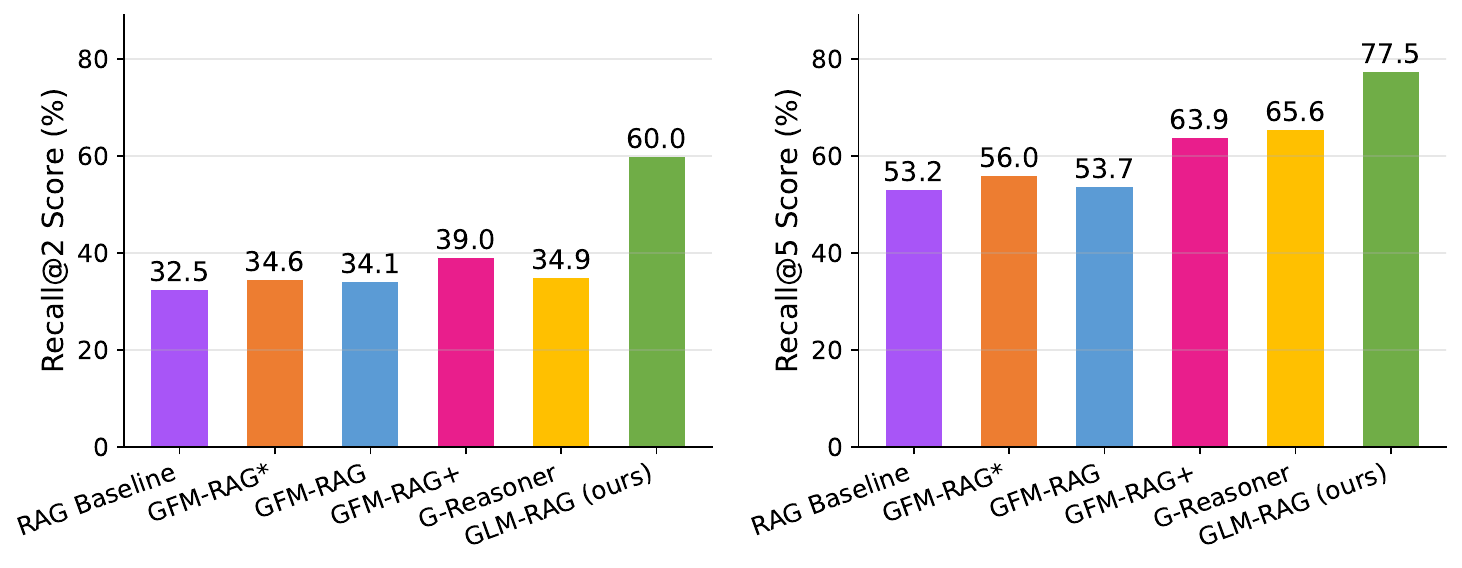}
    \caption{Retrieval performance (Recall@2 and Recall@5) on MultiHopRAG, an out-of-domain dataset with multi-hop questions.}
    \label{fig:ext-retrieval-multihop}
\end{figure*}

\Cref{fig:ext-retrieval,fig:ext-retrieval-transfer,fig:ext-retrieval-multihop} show the Recall@2 and Recall@5 scores of each model. Additionally, \Cref{tab:wiki_recall_splits} shows a breakdown of the retrieval performance under the low-data and full-data settings.

There were several issues with the originally reported results of GFM-RAG. The results in the initial paper \citep{main} were affected by a bug in the code. Later, the authors reported updated results of GFM-RAG in their follow-up paper \cite{greasoner}. However, in this version they evaluated HotpotQA on a different test set. To avoid confusion arising from these inconsistencies, we report the reproduced results of GFM-RAG in our main paper, and \Cref{tab:retrieval,tab:qa_results} compare the originally reported and reproduced results of GFM-RAG. As shown in \Cref{tab:retrieval}, the only notable difference appears, as expected, on HotpotQA.

\Cref{tab:retrieval} also shows retrieval comparisons with previous works. The methods are categorized as \textit{``non-structure methods''}, meaning that they do not utilize graph-based approaches, and \textit{``graph-enhanced methods''}. We would like to highlight again that integrating our proposed GLM retriever into the current SOTA method, G-Reasoner, is feasible, since it currently employs a GNN-based retriever, and leave this integration to future work.

\subsection{Retrieval Performance by Question Complexity }\label{app:perhop}

Furthermore, we wanted to analyze the retrieval performance of each model across different levels of question complexity. Thus, we examined retrieval success at both grouping the questions' complexity at their document and entity level.

\subsubsection*{Document Level}

\begin{figure*}[b]
    \centering
    \includegraphics[width=\linewidth]{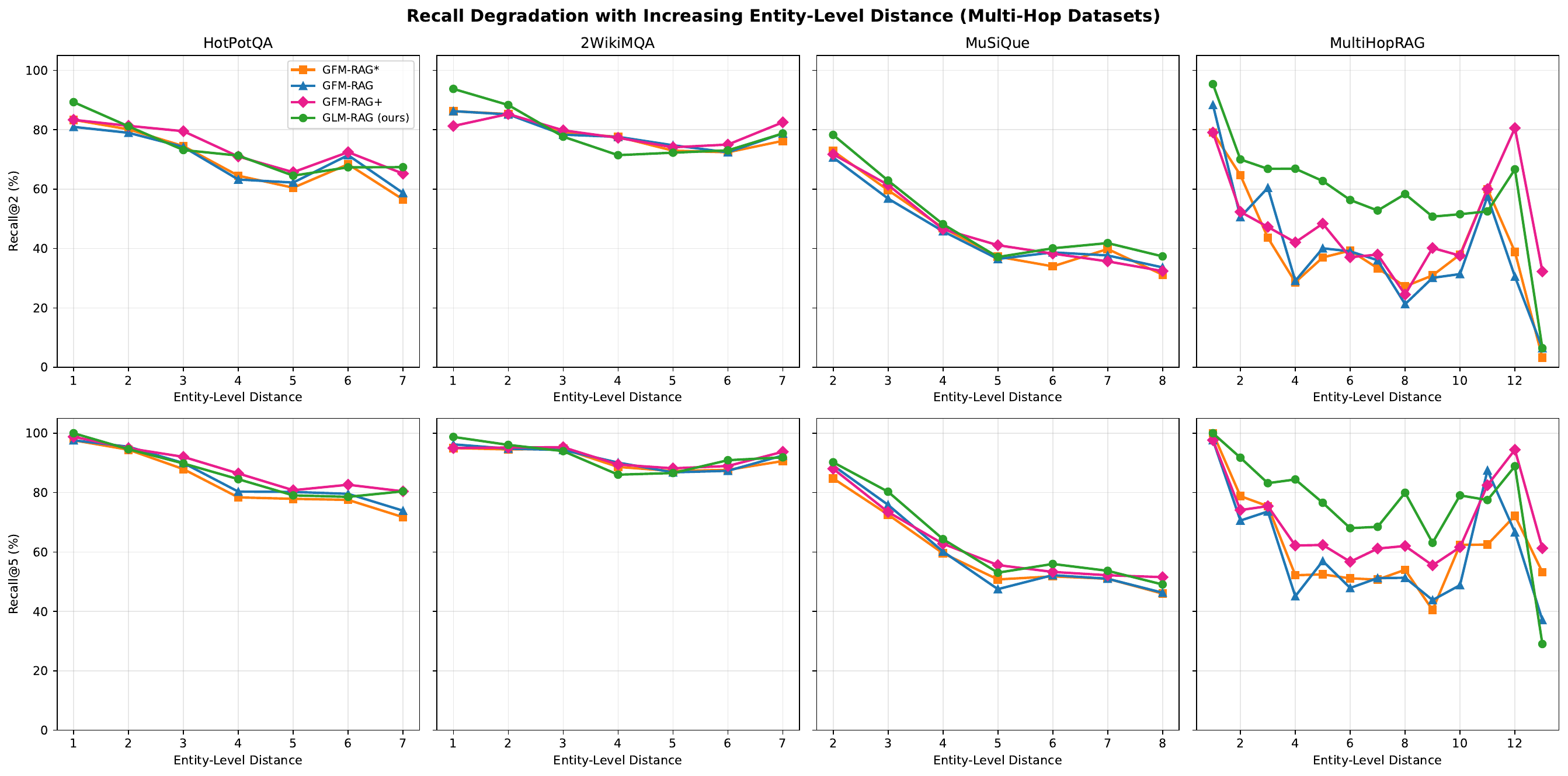}
    \caption{Recall degradation with increasing entity-level distance between golden and retrieved entities for multi-hop datasets.}
    \label{fig:recall_degradation_doc_multi}
\end{figure*}

\begin{figure*}[b]
    \centering
    \includegraphics[width=\linewidth]{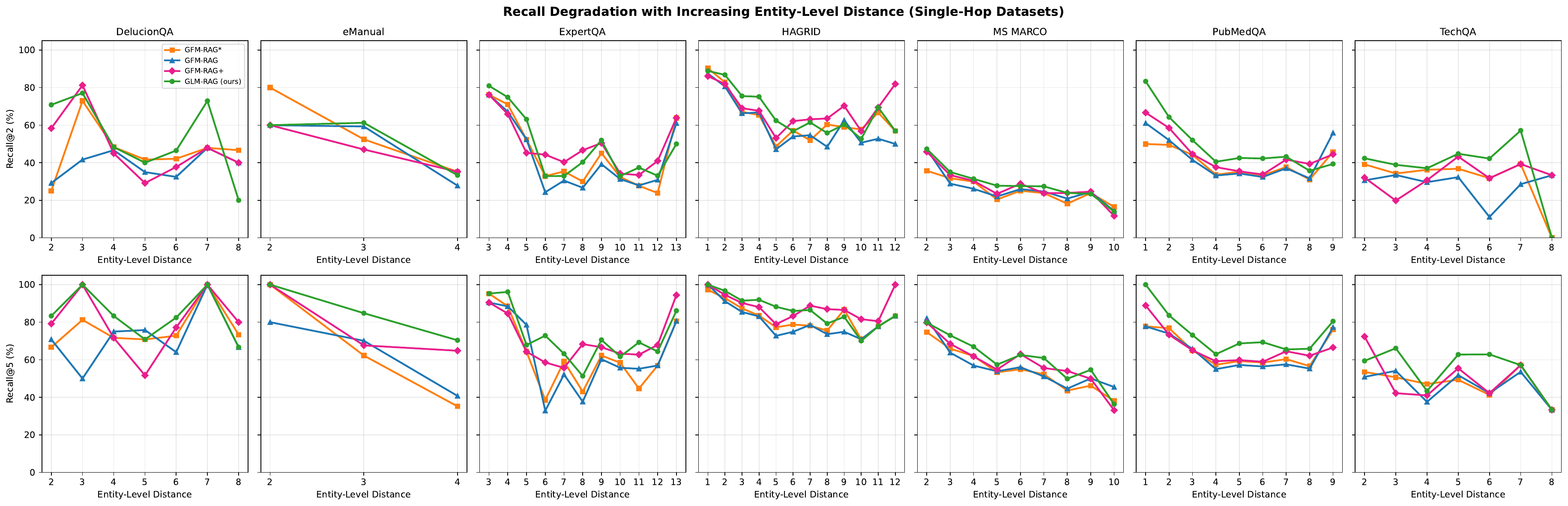}
    \caption{Recall degradation with increasing entity-level distance between golden and retrieved entities for single-hop datasets.}
    \label{fig:recall_degradation_doc_single}
\end{figure*}

To evaluate retrieval success at the document level, we grouped questions by the number of supporting documents they require. \Cref{fig:recall_degradation_doc_multi} shows the recall degradation with increasing numbers of supporting documents for the multi-hop datasets, while \Cref{fig:recall_degradation_doc_single} shows the same analysis for the single-hop datasets.

\subsubsection*{Entity Level} 

Our entity-level measure of question complexity captures how far the hardest-to-reach supporting entity lies from the question entities in the KG.

As shown in \Cref{lst:json-example}, there are almost always multiple question (seed) entities as well as multiple golden (supporting) entities for a question. Therefore, we calculate the distances from all question entities to all supporting entities and use the maximum of these distances for each question.

Let $G = (V, E)$ be the undirected KG, $S = \{s_1, \ldots, s_m\} \subseteq V$ the set of question entities, and $T = \{t_1, \ldots, t_n\} \subseteq V \setminus S$ the set of supporting entities. For each supporting entity $t_j$, we compute its minimum shortest-path distance to any seed entity:

$$
d(t_j, S) = \min_{s \in S} \, d_G(s, t_j)
$$

The entity-level distance is then defined as the maximum over all reachable supporting entities:

$$
\text{depth}(S, T) = \max_{\substack{j=1,\ldots,n \\ d(t_j, S) \neq \infty}} d(t_j, S)
$$

This captures the minimum number of relational hops the model must traverse to reach the most distant piece of evidence. Questions are grouped according to this depth value to analyze how retrieval performance degrades as reasoning complexity increases.

It is important to note that the entity-level distances in the ``single-hop datasets'' do not imply that these are ``multi-hop datasets'', since these hops do not cross document boundaries.

\begin{figure*}[b]
    \centering
    \includegraphics[width=\linewidth]{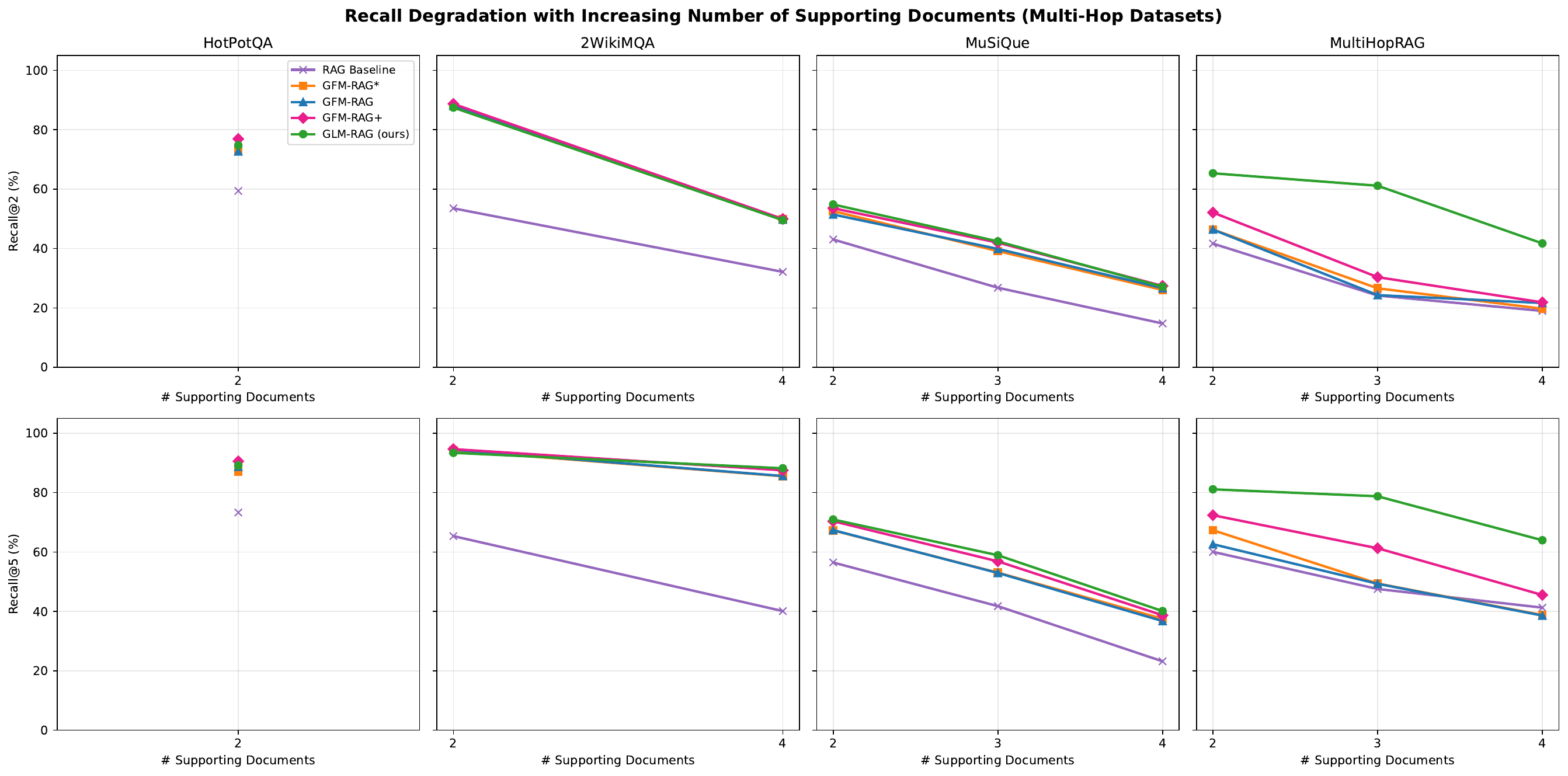}
    \caption{Recall degradation with increasing number of supporting documents for multi-hop datasets.}    
    \label{fig:recall_degradation_entity_multi}
\end{figure*}

\begin{figure*}[b]
    \centering
    \includegraphics[width=\linewidth]{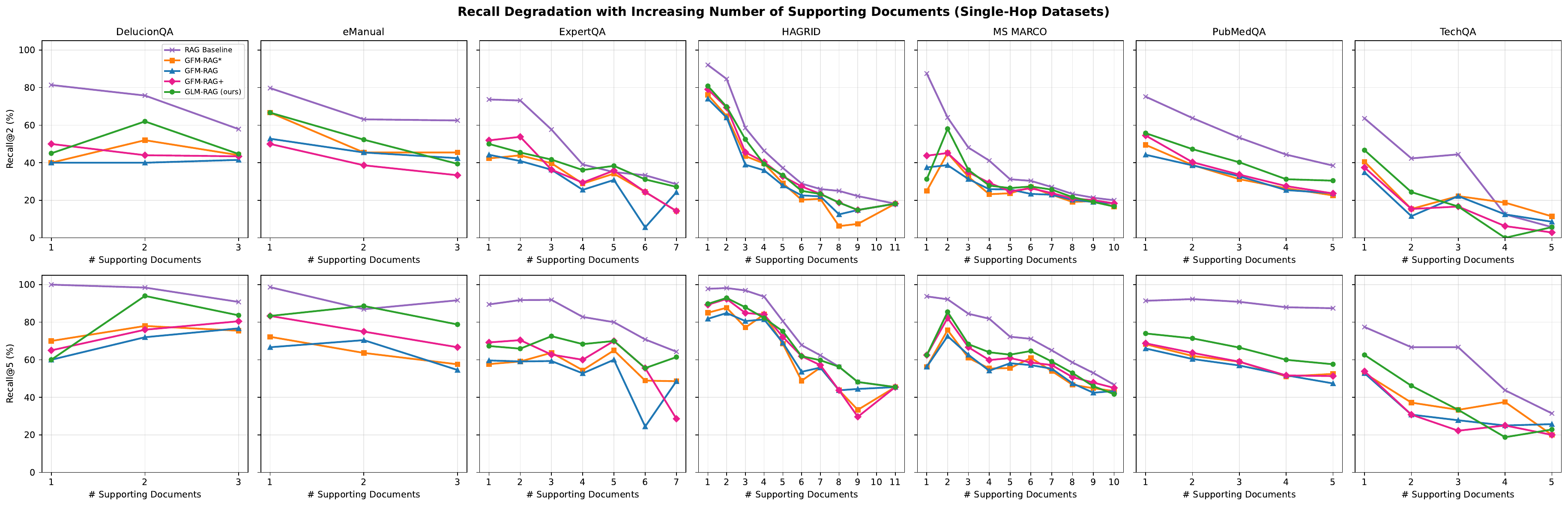}
    \caption{Recall degradation with increasing number of supporting documents for single-hop datasets.}
    \label{fig:recall_degradation_entity_single}
\end{figure*}

\Cref{fig:recall_degradation_entity_multi,fig:recall_degradation_entity_single} show the recall degradation with increasing entity-level distance for both single-hop and multi-hop datasets. All models follow a similar trend, with performance decreasing as the entity-level distance increases.

\begin{figure}[ht]
\centering
\begin{lstlisting}[basicstyle=\footnotesize\ttfamily, frame=single, framesep=2mm, backgroundcolor=\color{gray!10}, columns=fullflexible, breaklines=true]
{"id": "cc7f68a9086c11ebbd61ac1f6bf848b6",
"question": "Are The M Machine and 
Signals Midwest from the same country?",
"answer": "yes",
"answer_aliases": [],
"supporting_facts": 
    ["The M Machine",
    "Signals Midwest"],
"question_entities": 
    ["the m machine",
    "signals midwest"],
"supporting_entities": 
    ["ohio",
    "cleveland",
    "american punk rock band",
    "signals midwest",
    "2011",
    "san francisco",
    "united states",
    "american electronic music duo",
    "four eps",
    "skrillex s label owsla",
    "overall  1 slot on beatport",
    "the m machine",
    "ben swardlick",
    "two remix collections",
    "california",
    "eric luttrell",
    "a single"]
}
\end{lstlisting}
\caption{An example test question from 2Wiki.}
  \label{lst:json-example}
\end{figure}

\subsection{Retrieval Performance by Subgraph Budget}

In order to analyze the subgraph budget behavior on different neighborhoods, we separate questions into three categories, namely sparse, medium and dense, using the number of edges around the seed entities. The density distribution of the queries are more on the sparse side, however we account for this by separating the queries in equal terciles. Then we analyze whether the golden documents are reachable with our subgraph selection method under varying hop and triplet counts. This analyzes the \textbf{coverage} with increasing subgraph budget.

\begin{figure*}[b]
    \centering
    \includegraphics[width=\linewidth]{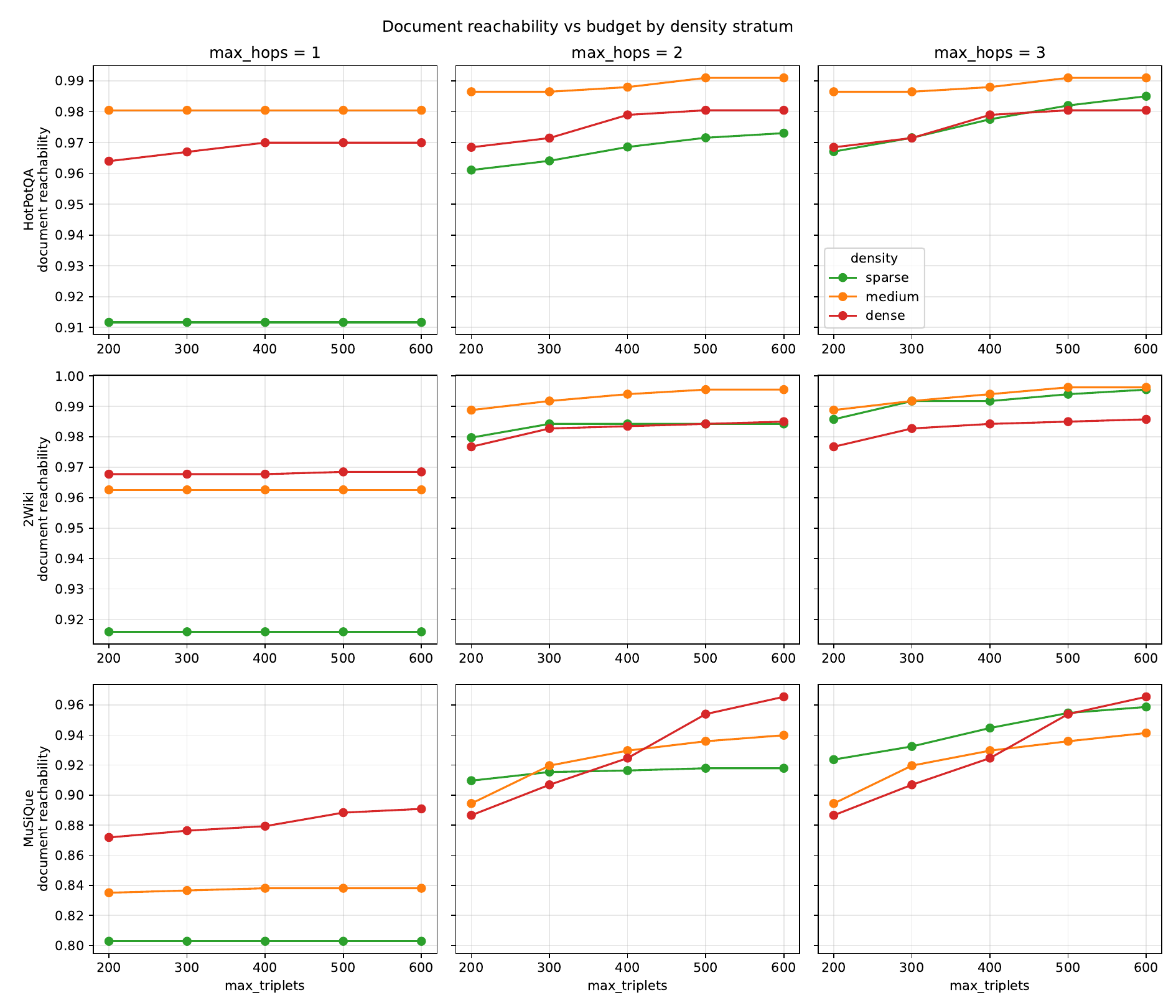}
    \caption{Golden document reachability of GLM-RAG split into 3 question categories regarding their neighborhood density with increasing number of hops and triplets.}
    \label{fig:budget_density}
\end{figure*}

\Cref{fig:budget_density} shows that in dense neighborhoods, 1-hop alone can already exceed the budget, so increasing max\_hops without increasing the triplet budget does not help. In sparse neighborhoods the opposite holds; extra hops help, the triplet count rarely binds. However, for instance on HotpotQA 2 hops and 200 triplets are enough to reach >96\% of the gold documents in all three density categories.

\begin{figure*}[b]
    \centering
    \includegraphics[width=0.8\linewidth]{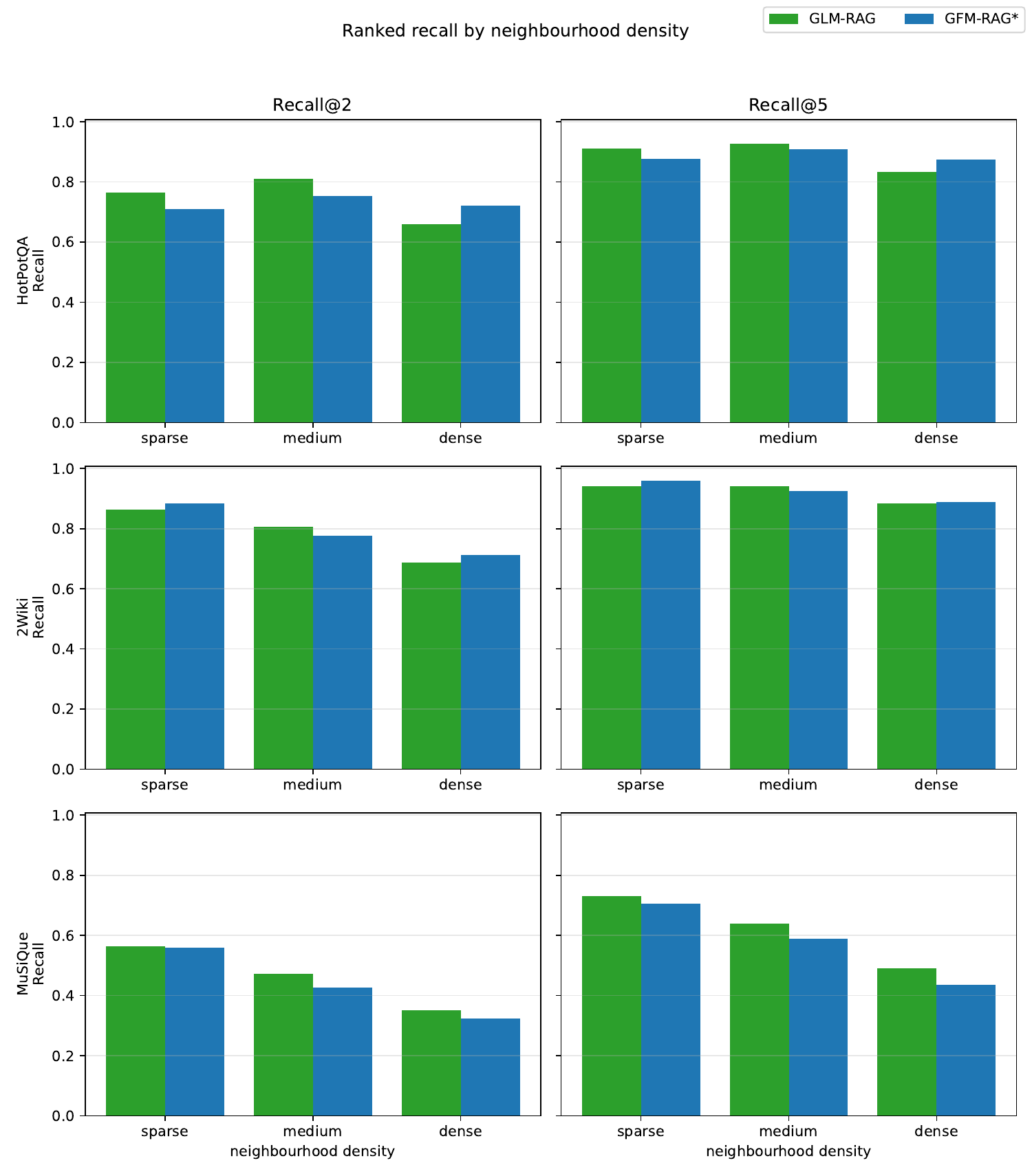}
    \caption{Retrieval recall of GLM-RAG and GFM-RAG* split into 3 question categories regarding their neighborhood density.}
    \label{fig:model_density}
\end{figure*}

Following these findings, we have also analyzed how the \textbf{retrieval performance} is affected from the density of the neighborhood. \Cref{fig:model_density} shows degradation of performance with higher density questions on 2Wiki and MuSiQue for both models, however the trend is not so clear on HotPotQA. Considering the visibility restriction of GLM-RAG and despite the full visibility of GFM-RAG*, we still see similar degrading trends in dense neighborhoods, meaning that our restriction does not behave disruptively worse in practice.

\section{Extensive QA Results}\label{app:qa}

\begin{figure*}[b]
    \centering
    \includegraphics[width=1\linewidth]{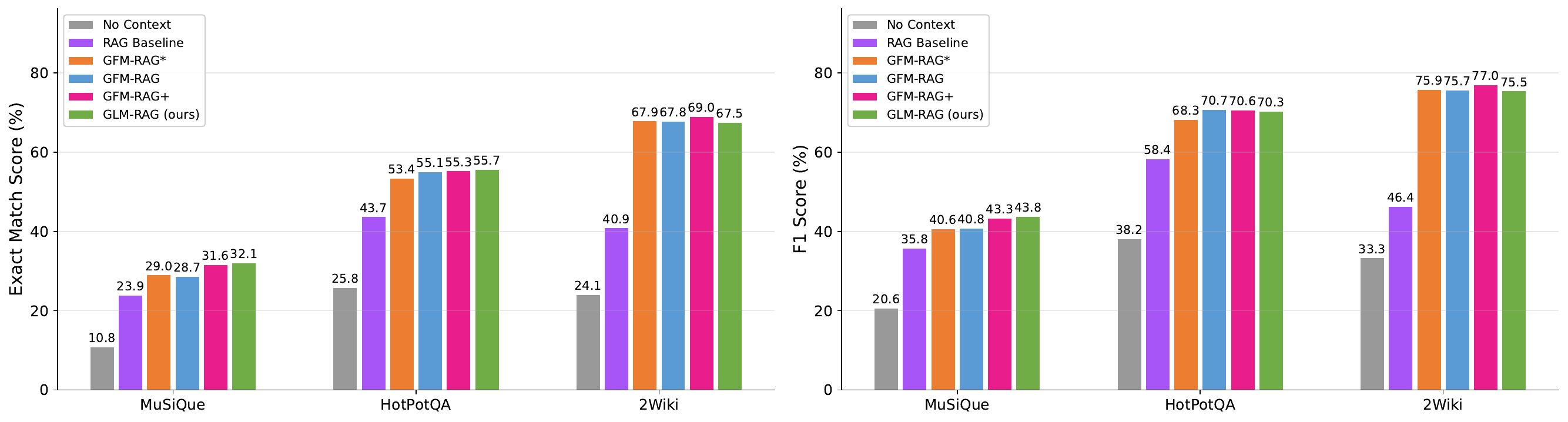}
    \caption{QA performance (Exact Match and F1) on in-domain Wikipedia datasets with \texttt{gpt-4o-mini}.}
    \label{fig:qa-extensive}
\end{figure*}

\Cref{fig:qa-extensive} shows the Exact Match (EM) and F1 scores for the Wikipedia datasets, and \Cref{tab:qa_results} provides comparisons with additional baselines.

\Cref{fig:prompt} shows the prompt example provided to the LLM.

\Cref{tab:qa-subgraph} shows an additional baseline where the model receives a subgraph instead of documents in its context. The subgraph selection is done via the same method in GLM-RAG, starting from the question entities and selecting 600 triplets in a 2-hop neighborhood.

In our RAG pipeline, the knowledge graph is constructed by extracting entities from documents, and the retriever acts on a subgraph to find the relevant entities. The entities are then mapped back to the documents so that the “relevant” documents can be used for generation. 

Although simply providing the entity-based knowledge subgraph would not be an apples-to-apples comparison and we expect it to underperform, we want to highlight the importance of giving the whole document as context instead of just the graphs itself since they contain more information. 

\Cref{tab:qa-subgraph} shows how the \textbf{subgraph baseline} compares against other methods on MuSiQuE, with \texttt{gpt-4o-mini} as the LLM. This resulted in a weaker baseline that underperforms other graph-based methods, showing the need for the full-context with documents in question answering.

\begin{table}[ht]
\centering
\caption{QA results with a subgraph baseline. Just giving the subgraph underperforms the document baselines.}
\label{tab:qa-subgraph}
\small
\begin{tabular}{lcc}
\toprule
\setlength{\tabcolsep}{4pt}
MuSiQuE  & EM & F1 \\ \midrule
No-Context Baseline & 10.8 & 20.6 \\
\textit{Subgraph Baseline} & \textit{22.9} & \textit{30.7} \\
RAG Baseline & 23.8 & 33.3 \\
GFM-RAG & 28.7 & 40.8 \\
GFM-RAG+ & 31.6 & 43.3 \\
GFM-RAG* & 29.0 & 40.6 \\
GLM-RAG & \textbf{32.1} & \textbf{43.8} \\
\bottomrule
\end{tabular}
\end{table}

\section{Extensive Transferability Results}\label{app:multihop}

\Cref{tab:transferability_qa} shows the models' F1 scores on the downstream QA task. There are two reasons why the high retrieval performance is not fully reflected in the QA results:

Firstly, in comparison to the Wiki datasets, the golden answers for these transfer datasets are full sentences rather than single words or phrases. This means that QA metrics such as EM and F1 do not fully capture the quality of the generated answers.

Secondly, the strong performance of the no-context baseline indicates that much of this information is already contained within \texttt{gpt-4o-mini}'s parametric knowledge, meaning that the model does not necessarily require the retrieved context to answer the questions correctly. This potentially dilutes the impact a strong retrieval can have. 

Nonetheless, GLM-RAG achieves the best or second-best F1 scores on 7 out of 8 datasets; however, the performance gaps remain relatively small due to the issues described above.

The same problem arises in the QA evaluation of MultihopRAG. Given that \texttt{gpt-4o-mini} can already achieve quite high scores without any additional context, shown on the no context baseline on \Cref{tab:multi-qa}, we conclude that both the exact match and the accuracy measures are not expressive of the retrieval quality.

\begin{table}[ht]
\centering
\caption{Retrieval (Recall@2 and Evidence Recall from G-Bench \cite{whengraph}) and QA performance (EM and ACC from G-Bench \cite{whengraph}) on MultihopRAG, answers generated with \texttt{gpt-4o-mini}.}
\label{tab:multi-qa}
\small
\begin{tabular}{lcccc}
\toprule
 & R@2 & RE & EM & ACC \\
\midrule
No Context & - & - & 72.8  & 75.7  \\
RAG Baseline & 32.5 & 55.8 & \textbf{79.4} & \textbf{78.1}  \\
GFM-RAG & 34.1 & 54.5 & 71.1 & 73.5  \\
GFM-RAG+ & 39.0 & 55.8 & 72.7 & 74.2  \\
GLM-RAG (ours) & \textbf{60.0} & \textbf{58.1} & \underline{75.5} & \underline{77.1}  \\
\bottomrule
\end{tabular}
\end{table}

\section{Extensive Analysis}\label{app:ext-analysis}

In the following, we explain how the similarity and distance measures (see \Cref{sec:analysis}) are computed. \Cref{tab:semantic_sim_comparison,tab:mean_hop_dist} show the overview the models' similarity and distance measures.

\subsection{Similarity Measures}

\begin{table}[ht]
\centering
\caption{Comparison of cosine similarity between the question and the retrieved entities.}
\label{tab:semantic_sim_comparison_2}
\resizebox{\columnwidth}{!}{%
\begin{tabular}{l|c|ccc}
\toprule
\textbf{Models} & GLM-RAG & GFM-RAG*  & GFM-RAG & GFM-RAG+ \\
\midrule
\multicolumn{5}{c}{\cellcolor[HTML]{C0C0C0}\textbf{Multi-Hop Datasets}} \\ \midrule
MuSiQue & 0.309 & 0.283 & 0.286 & 0.297 \\
HotPotQA & 0.312 & 0.292 & 0.293 & 0.298 \\
2Wiki & 0.285 & 0.262 & 0.265 & 0.268 \\
MultiHopRAG & 0.245 & 0.209 & 0.212 & 0.241 \\
\midrule
\multicolumn{5}{c}{\cellcolor[HTML]{C0C0C0}\textbf{Single-Hop Datasets}} \\ \midrule
TechQA & 0.255 & 0.234 & 0.228 & 0.245 \\
PubMedQA & 0.336 & 0.298 & 0.289 & 0.314 \\
MS MARCO & 0.294 & 0.280 & 0.280 & 0.314 \\
HAGRID & 0.345 & 0.329 & 0.323 & 0.355 \\
ExpertQA & 0.284 & 0.249 & 0.235 & 0.286 \\
eManual & 0.276 & 0.234 & 0.221 & 0.286 \\
DelucionQA & 0.294 & 0.287 & 0.286 & 0.313 \\
\bottomrule
\end{tabular}
}
\end{table}

Semantic similarity between the question and retrieved entities is computed as the cosine similarity of their sentence-level embeddings. Specifically, we encode the question text $q$ and each retrieved entity name $e_i$ using a pretrained sentence embedding model (\texttt{all-mpnet-base-v2}), yielding embedding vectors $\mathbf{q} = \mathrm{Enc}(q)$ and $\mathbf{e}_i = \mathrm{Enc}(e_i)$. Each embedding is $\ell_2$-normalized:

$$
\hat{\mathbf{q}} = \frac{\mathbf{q}}{\|\mathbf{q}\|_2},
\quad
\hat{\mathbf{e}}_i = \frac{\mathbf{e}_i}{\|\mathbf{e}_i\|_2}
$$

The semantic similarity score for a retrieved entity $e_i$ with respect to the question $q$ is then defined as:

$$
\mathrm{sim}(q, e_i) = \hat{\mathbf{q}}^\top \hat{\mathbf{e}}_i
$$

which is equivalent to the cosine similarity $\cos(\mathbf{q}, \mathbf{e}_i) \in [-1, 1]$. For each sample, the scores are computed over the top-$k$ retrieved entities and averaged to produce a per-sample semantic relevance score.

\subsection{Distance Measures}

\begin{table}[ht]
\centering
\caption{Comparision of mean hop distance between the seed entities and the retrieved entities.}
\label{tab:mean_hop_dist}
\resizebox{\columnwidth}{!}{%
\begin{tabular}{l|c|ccc}
\toprule
\textbf{Models} & GLM-RAG & GFM-RAG* & GFM-RAG & GFM-RAG+ \\
\midrule
\multicolumn{5}{c}{\cellcolor[HTML]{C0C0C0}\textbf{Multi-Hop Datasets}} \\ \midrule
MuSiQue & 1.09 & 1.22 & 1.29 & 1.34 \\
HotpotQA & 1.05 & 1.19 & 1.23 & 1.24 \\
2Wiki & 1.10 & 1.19 & 1.22 & 1.24 \\
MultiHopRAG & 0.88 & 1.02 & 0.99 & 1.24 \\
\midrule
\multicolumn{5}{c}{\cellcolor[HTML]{C0C0C0}\textbf{Single-Hop Datasets}} \\ \midrule
TechQA & 1.06 & 1.05 & 1.17 & 1.10 \\
PubMed & 1.04 & 1.19 & 1.31 & 1.44 \\
MS MARCO & 1.25 & 1.26 & 1.39 & 1.49 \\
HAGRID & 1.32 & 1.32 & 1.42 & 1.56 \\
ExpertQA & 1.53 & 1.59 & 1.89 & 2.06 \\
eManual & 1.29 & 1.38 & 1.50 & 1.55 \\
DelucionQA & 1.31 & 1.35 & 1.56 & 1.49 \\
\bottomrule
\end{tabular}
}
\end{table}

Structural proximity between retrieved entities and the question (seed) entities is measured as the shortest-path distance in the knowledge graph. Let $G = (V, E)$ be the undirected knowledge graph and $S = \{s_1, \ldots, s_m\} \subseteq V$ the set of seed entities extracted from the question. For each retrieved entity $e_i$, we compute its minimum shortest-path distance to any seed entity:

$$
d(e_i, S) = \min_{s \in S} \, d_G(s, e_i)
$$

where $d_G(s, e_i)$ is the length of the shortest path between $s$ and $e_i$ in $G$, obtained via breadth-first search with a cutoff of 10 hops. If no path exists within the cutoff, the entity is considered unreachable. The per-sample structural proximity score is defined as the average distance over all reachable top-$k$ retrieved entities:

$$
\bar{d} =
\frac{1}{|\{i : d(e_i, S) \neq \infty\}|}
\sum_{\substack{i=1 \\ d(e_i,S)\neq\infty}}^{k}
d(e_i, S)
$$

Lower values indicate that the retrieval model favors entities that are structurally close to the question entities in the KG.

\section{Error Bars and Significancy Tests}\label{app:significancy}

\Cref{tab:error} shows significance tests with paired bootstrap tests and 95\% confidence intervals (CI) that compare GLM-RAG against all other baselines. Holm-Bonferroni correction is applied to account for family-wide error rates. Significant results are indicated with one or multiple stars \textbf{(*)} indicating that p-values are; p: *** <.001  ** <.01  * <.05  ns >=.05

The results confirm that although the differences between GLM-RAG and GFM-RAG+ are not significant in in-domain retrieval, they are significant in out-of-domain retrieval.

\section{Declaration of AI usage}
We use AI assistants for speeding up programming, and to aid with reformulations. The content of this work is our own, and not inspired by AI assistants. 

\begin{table*}[b]
\centering
\caption{Statistics of the datasets and constructed KG-indexes used for \textbf{training}.}
\label{tab:train_dataset}
\resizebox{0.9\linewidth}{!}{%
\begin{tabular}{cccccc}
\toprule
Dataset & \# Queries & \# Documents & \# Entities & \# Relations & \# Triplets \\
\midrule
HotpotQA (train) & 89,447 & 874,784 & 8,259,397 & 4,134,914 & 27,362,937 \\
HotpotQA (valid) & 1,000 & 9,742 & 92,651 & 46,290 & 314,944 \\
MuSiQue (train) & 18,938 & 378,724 & 1,374,033 & 801,704 & 4,315,300 \\ 
MuSiQue (valid) & 1,000 & 20,000 & 70,080 & 42,690 & 213,797 \\
2WikiMultihopQA (train) & 165,454 & 955,412 & 7,120,577 & 2,876,828 & 22,188,618 \\
2WikiMultihopQA (valid) & 1,000 & 5,761 & 42,489 & 17,228 & 130,397 \\
\midrule
Total Train & 273,830 & 2,208,920 & 18,579,900 & 3,814,333 & 53,866,855 \\
\bottomrule
\end{tabular}
}
\label{tab:dataset_stats}
\end{table*}

\begin{table*}[b]
\centering
\caption{Statistics of the datasets and constructed KG-indexes used for \textbf{testing}.}
\label{tab:test_dataset}
\resizebox{.9\linewidth}{!}{%
\begin{tabular}{ccccccc}
\toprule
Dataset    & Domain            & \#Queries & \#Documents & \#Entities & \#Relations & \#Triplets \\ \midrule
HotpotQA (test)  & Wikipedia         & 1,000 & 9,221     & 87,768  & 45,112    & 279,112 \\
MuSiQue (test)   & Wikipedia         & 1,000 & 11,656    & 100,853 & 55,944    & 319,618 \\
2Wiki (test)     & Wikipedia         & 1,000 & 6,119     & 48,779  & 20,748    & 160,950 \\
MultiHopRAG & News             & 2,255 & 609       & 16,147  & 9,416     & 24,180 \\
PubMedQA   & Biomedical        & 2,450 & 5,932     & 42,389  & 20,952    & 149,782 \\
DelucionQA & Customer Support  & 184   & 235       & 2,669   & 2,298     & 6,183   \\
TechQA     & Customer Support  & 314   & 769       & 10,221  & 4,606     & 57,613  \\
ExpertQA   & Customer Support  & 203   & 808       & 11,079  & 6,810     & 16,541  \\
EManual    & Customer Support  & 132   & 102       & 695     & 586       & 1,329   \\
MS Marco   & General Knowledge & 423   & 3,481     & 24,740  & 17,042    & 63,995  \\
HAGRID     & General Knowledge & 1,318 & 1,975     & 23,484  & 18,653    & 48,969  \\ 
G-Bench (CS) & Computer Science & 1,018 & 15,011 & 187,217 & 79,578 & 991,630 \\
G-Bench (Novel) & Novels & 2,010 & 1,881 & 39,870 & 37,344 & 68,483 \\
G-Bench (Medical) & Medical & 2,062 & 390 & 8,676 & 6,814 & 32,386 \\ \bottomrule
\end{tabular}%
}
\end{table*}

\begin{table*}[b]
\centering
\caption{Graph statistics of the datasets and constructed KG-indexes used for \textbf{training}.}
\label{tab:train_graph}
\resizebox{.8\linewidth}{!}{%
\begin{tabular}{ccccc}
\toprule
Dataset & Avg Degree & Density & \# Components & Largest CC \%  \\
\midrule
HotpotQA (train) & 5.89 & 0.0001 & 138 & 99.66 \\
HotpotQA (valid) & 6.02 & 0.0001 & 125 & 99.69 \\
MuSiQue (train) & 5.59 & 0.0001 & 140 & 99.53 \\
MuSiQue (valid) & 5.48 & 0.0001 & 137 & 99.55 \\
2Wiki (train) & 5.47 & 0.0001 & 92 & 99.50  \\
2Wiki (valid) & 5.40 & 0.0001 & 75 & 99.59 \\
\bottomrule
\end{tabular}
}
\label{tab:graph_stats}
\end{table*}

\begin{table*}[b]
\centering
\caption{Graph statistics of the test datasets and constructed KG-indexes used for \textbf{testing}.}
\label{tab:test_graph}
\resizebox{.8\linewidth}{!}{%
\begin{tabular}{ccccc}
\toprule
Dataset & Avg Degree & Density & \# Components & Largest CC \% \\
\midrule
HotpotQA (test) & 5.68 & 0.0001 & 150 & 99.60 \\
MuSiQue (test) & 5.70 & 0.0001 & 201 & 99.56 \\
2Wiki (test) & 5.77 & 0.0001 & 120 & 99.41 \\
MultiHopRAG & 2.84 & 0.0002 & 378 & 93.63\\
PubMedQA & 6.28 & 0.0001 & 129 & 99.27 \\
DelucionQA & 4.11 & 0.0015 & 31 & 97.45\\
TechQA & 8.98 & 0.0009 & 98 & 97.36 \\
ExpertQA & 2.77 & 0.0002 & 449 & 89.99 \\
EManual & 3.42 & 0.0049 & 18 & 94.39 \\
MS Marco & 4.62 & 0.0002 & 124 & 98.82  \\
HAGRID & 3.81 & 0.0002 & 253 & 97.26 \\
G-Bench (CS) & 9.60 & 0.0001 & 558 & 99.29 \\
G-Bench (Novel) & 3.17 & 0.0020 & 34.25 & 95.63 \\ %
G-Bench (Medical) & 6.60 & 0.0008 & 39 & 98.80 \\ 
\bottomrule
\end{tabular}
}
\end{table*}

\begin{table*}[b]
\centering
\caption{The detailed implementation and training settings of GFM-RAG, GFM-RAG*, GFM-RAG+ and GLM.}
\label{tab:settings}
\resizebox{0.8\linewidth}{!}{%
\begin{tabular}{ccc}
\toprule
Setting & \textbf{GFM-RAG (*, +)} & \textbf{GLM-RAG}          \\ \midrule
\multicolumn{3}{c}{\cellcolor[HTML]{C0C0C0}\textbf{KG-index Construction}} \\ \midrule

OpenIE                   & GPT-4o-mini &   GPT-4o-mini      \\
Entity resolution        & ColBERTv2 &   ColBERTv2           \\ \midrule
\multicolumn{3}{c}{\cellcolor[HTML]{C0C0C0}\textbf{Models}} \\ \midrule
 Backbone   &     GNN (based on \citet{ultra}) & GLM (based on T5-large)  \\                 
 \# (Encoder) Layer                  & 6     & 12               \\
\# Parameters & 8,144,897 & 336M \\
Hidden dim   (($d_\text{model}$))            & 512    & 1024               \\
Attention heads  & -       & 16                   \\
Graph encoding   & -          & Levi graph + global RPE \\
Message                  & DistMult   & -          \\
Aggregation              & Sum      & -            \\
Entity scorer           & 2-layer MLP & 2-layer MLP          \\
Sentence embedding model & all-mpnet-base-v2 &   all-mpnet-base-v2       \\ 
Doc ranker entities & 20 & 20                  \\\midrule
\multicolumn{3}{c}{\cellcolor[HTML]{C0C0C0}\textbf{Finetuning}} \\ \midrule
$\alpha$            & 0.3    &     0.44       \\
Optimizer               & AdamW & AdamW             \\
(Head) Learning rate            & 5e-4  &   5e-4         \\
Backbone learning rate  & -   & 1e-4                 \\
Batch size               & 4    & 2                \\
Training epochs          & 5     & 2               \\
Max triplets    & -           & 600                  \\ 
\# Hops         & -         & 2 \\
                                 \bottomrule

\end{tabular}%
 }
\end{table*}

\begin{table*}[b]
\label{tab:ablation}
\centering
\caption{Ablation study results averaged over HotpotQA, MuSiQue, and 2Wiki test sets.}
\label{tab:ablation}
\begin{tabular}{llcc}
\toprule
Ablation & Method & Recall@2 & Recall@5 \\
\midrule
Embedding Aggregation & Mean (default) & 63.3 & 79.6 \\
 & First & 63.4 & 79.6 \\
 & Last & 63.3 & 79.2 \\
\midrule
Query Interaction \\ \small{[yes/no and use\_text=FullyConnected/no]} & Yes and FC (default) & 63.3 & 79.6 \\
 & No and FC & 63.5 & 79.4 \\
 & Yes and No FC & 58.7 & 74.6 \\
 & No and No FC & 57.3 & 73.4 \\
\midrule
Number of Hops & 1 Hop & 49.4 & 66.4 \\
 & 2 Hops (default) & 63.3 & 79.6 \\
 & 3 Hops & 63.0 & 79.7 \\
 & 4 Hops & 63.6 & 79.6 \\
\midrule
Max Triplets & 300 & 60.8 & 76.9 \\
 & 400 & 61.8 & 78.4 \\
 & 500 & 62.8 & 79.1 \\
 & 600 (default) & 63.3 & 79.6 \\
\midrule
T5 Encoder Size & T5-Small (60M) & 60.4 & 76.7 \\
& T5-Base (220M) (default) & 63.3 & 79.6 \\
 & T5-Large (770M) & 64.7 & 80.6 \\
\midrule
Graph Aware Attention \small max\_triplets = 300 \normalsize & With & 60.8 & 76.9 \\
& Without (linearized) & 38.7 & 51.3 \\
\bottomrule
\end{tabular}
\end{table*}

\begin{table*}[b]
\centering
\caption{Retrieval performance (Recall@2 and Recall@5) on in-domain Wikipedia datasets in low data and full data settings of comparable graph methods. Best results are \textbf{bolded} and second best are \underline{underlined}.}
\label{tab:wiki_recall_splits}
\begin{tabular}{lccccccc}
\toprule
 & & \multicolumn{2}{c}{HotpotQA} & \multicolumn{2}{c}{MuSiQue} & \multicolumn{2}{c}{2Wiki} \\ \cmidrule(l){3-8} 
Data Amount & Method & R@2 & R@5 & R@2 & R@5 & R@2 & R@5 \\ \midrule
\multirow{4}{*}{60k Q-Doc Pairs} & RAG Baseline & 48.5 & 46.3 & 33.2 & 59.4 & 73.3 & 59.4 \\ 
\midrule
& GFM-RAG (reproduced) &  69.5 & 85.9 & 42.8 & 57.1 & \underline{77.4} & 91.5 \\
& GFM-RAG* & 70.4 & 85.0 & 43.2 & 58.2 & 76.9 & 91.4 \\
& GFM-RAG+ & \textbf{72.7} & \textbf{88.8} & \underline{44.1} & \underline{59.1} & \textbf{77.4} & \textbf{92.2} \\
& GLM-RAG (ours) & \underline{71.2} & \underline{88.1} & \textbf{46.0} & \textbf{61.5} & 76.8 & \underline{92.1} \\
\midrule \midrule
\multirow{3}{*}{277k Q-Doc Pairs}
& GFM-RAG (reproduced) & 72.7 & 88.7 & 43.7 & 57.7 & \underline{79.0} & \underline{92.4} \\
& GFM-RAG* & 73.1 & 87.1 & 43.9 & 57.8 & \underline{79.0} & 92.2 \\
& GFM-RAG+ & \textbf{76.9} & \textbf{90.6} & \underline{45.6} & \underline{60.8} & \textbf{79.6} & \textbf{93.0} \\
& GLM-RAG (ours) & \underline{74.8} & \underline{89.0} & \textbf{46.4} & \textbf{62.2} & 78.6 & 92.2 \\
\bottomrule
\end{tabular}
\end{table*}

\begin{table*}[b]
\centering
\caption{Retrieval performance comparison (Recall@2 and Recall@5). Best results are \textbf{bolded}, second best are \underline{underlined}, and third best are \textit{italized}. Baseline scores are adapted from \citet{greasoner}.}
\label{tab:retrieval}
\begin{tabular}{lcccccc}
\toprule
 & \multicolumn{2}{c}{HotpotQA} & \multicolumn{2}{c}{MuSiQue} & \multicolumn{2}{c}{2Wiki} \\ \cmidrule(l){2-7} 
Method & R@2 & R@5 & R@2 & R@5 & R@2 & R@5 \\ \midrule
\multicolumn{7}{c}{\cellcolor[HTML]{C0C0C0}Non-structure Methods} \\ \midrule
BM25 \citep{bm25} & 55.4 & 72.2 & 32.3 & 41.2 & 51.8 & 61.9 \\
ColBERTv2 \citep{colbertv2} & 64.7 & 79.3 & 37.9 & 49.2 & 59.2 & 68.2 \\
Qwen3-Emb (8B) \citep{qwen3} & 74.1 & 88.8 & \textit{46.8} & 62.1 & 66.2 & 74.1 \\ \midrule
\multicolumn{7}{c}{\cellcolor[HTML]{C0C0C0}Graph-enhanced Methods} \\ \midrule
RAPTOR \citep{raptor} & 58.1 & 71.2 & 35.7 & 45.3 & 46.3 & 53.8 \\
GraphRAG (MS) \citep{graphrag} & 58.3 & 76.6 & 35.4 & 49.3 & 61.6 & 77.3 \\
LightRAG \citep{lightrag} & 38.8 & 54.7 & 24.8 & 34.7 & 45.1 & 59.1 \\
KAG \citep{KAG} & 59.4 & 86.1 & 42.2 & \textit{62.4} & 61.4 & 88.3 \\
HippoRAG \citep{hipporag} & 60.1 & 78.5 & 41.2 & 53.2 & 68.4 & 87.0 \\
HippoRAG 2 \citep{hipporag2} & \underline{80.5} & \underline{95.7} & \underline{53.5} & \underline{74.2} & \underline{80.5} & \underline{95.7} \\
SubgraphRAG \citep{subgraphrag} & 58.1 & 71.7 & 40.6 & 48.1 & 70.2 & 85.3 \\ 
\midrule
GFM-RAG \citep{main} & 75.6 & 89.6 & 43.5 & 57.6 & 79.1 & 92.4 \\
GFM-RAG (reproduced) & 72.7 & 88.7 & 43.7 & 57.7 & 79.0 & 92.4 \\
GFM-RAG* & 73.1 & 87.1 & 43.9 & 57.8 & 79.0 & 92.2 \\
GFM-RAG+ & \textit{76.9} & \textit{90.6} & 45.6 & 60.8 & \textit{79.6} & \textit{93.0} \\
GLM-RAG (ours) & 74.8 & 89.0 & 46.4 & 62.2 & 78.6 & 92.2 \\ \midrule
G-Reasoner \cite{greasoner} & \textbf{85.9} & \textbf{97.7} & \textbf{54.8} & \textbf{74.9} & \textbf{81.2} & \textbf{98.2} \\
\bottomrule
\end{tabular}
\end{table*}

\begin{table*}[b]
\centering
\caption{GNN-based and GLM-based retriever models parameter comparison when scaling up the GFM-RAG* by increasing its hidden dimensions from 512 to \textbf{1024}, \textbf{2048} and \textbf{4096}, following \citet{greasoner}. Since we only use the encoder, the parameter count of GLMs is half the original parameter count of T5-models.}
\label{tab:parameter-count}
\begin{tabular}{lcc}
\toprule
\setlength{\tabcolsep}{4pt}
Models & Parameter Count \\ \midrule
GLM-RAG (t5-small) & 36M \\
GFM-RAG(*, +) (hidden size 1024) & 31M \\ \midrule
GLM-RAG (t5-base) & 110M \\
GFM-RAG(*, +) (hidden size 2048) & 122M \\ \midrule
GLM-RAG (t5-large) & 336M \\
GFM-RAG(*, +) (hidden size 4096) & 476M \\ \bottomrule
\end{tabular}
\end{table*}

\begin{figure*}[b]
    \centering
    \includegraphics[width=\linewidth]{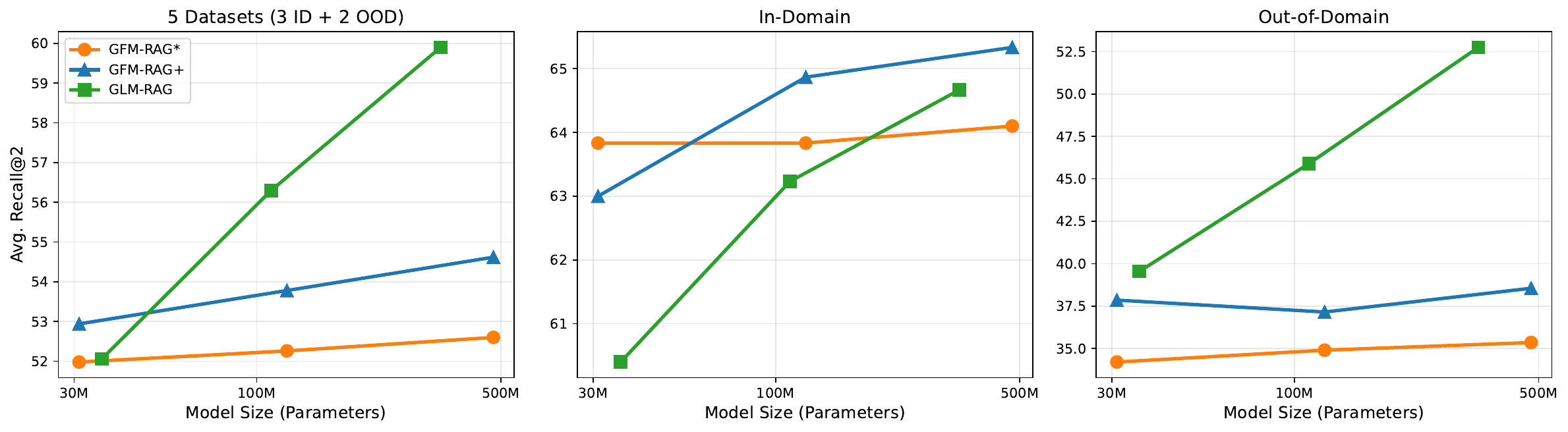}
    \caption{Comparision of GFM-RAG, GFM-RAG+ and GLM-RAG's retrieval quality on the three in-domain datasets (HotpotQA, MuSiQuE, 2Wiki), as well as one out-of-domain single-hop (PubMedQA) and one one out-of-domain multi-hop (MultiHopRAG). Scaling up the GFM-RAG variants doesn't bring the same contributions as scaling up the GLM-RAG in out-of-domain settings.}
    \label{fig:gnn-vs-glm-scale}
\end{figure*}

\begin{table*}[b]
\centering
\caption{Retrieval performance (Recall@2 and Recall@5) of the scaling experiments on the 60k dataset. Best results are \textbf{bolded}.}
\label{tab:gnn-up}
\resizebox{\linewidth}{!}{%
\begin{tabular}{lcccccccccc}
\toprule
 & \multicolumn{2}{c}{HotpotQA} & \multicolumn{2}{c}{MuSiQue} & \multicolumn{2}{c}{2Wiki} & \multicolumn{2}{c}{PubMedQA} & \multicolumn{2}{c}{MultiHopRAG} \\ \cmidrule(l){2-3} \cmidrule(l){4-5} \cmidrule(l){6-7} \cmidrule(l){8-9} \cmidrule(l){10-11}
Method & R@2 & R@5 & R@2 & R@5 & R@2 & R@5 & R@2 & R@5 & R@2 & R@5 \\ \midrule
GFM-RAG* \small(hidden\_dim=512, 8M) \normalsize  & 
70.4 & 85.0 & 43.2 & 58.2 & 76.9 & 91.4 & 35.8 & 58.8 & 29.7 & 50.1 \\
GFM-RAG* \small (hidden\_dim=1024, 31M) \normalsize & 70.3 & 86.6 & 43.5 & 57.4 & 77.7 & 92.0 & 35.1 & 59.8 & 33.3 & 53.4 \\ 
GFM-RAG* \small (hidden\_dim=2048, 122M) \normalsize & 71.3 & 87.0 & 42.3 & 57.5 & 77.9 & 91.3 & 35.0 & 57.4 & 34.8 & 57.6 \\ 
GFM-RAG* \small (hidden\_dim=4096, 476M) \normalsize & 71.8 & 86.7 & 42.9 & 57.4 & 77.6 & 91.7 & 36.1 & 59.8 & 34.6 & 56.1 \\ \midrule
GFM-RAG+ \small(hidden\_dim=512, 8M) \normalsize & 72.7 & \textbf{88.8} & 44.1 & 59.1 & 77.4 & \textbf{92.2} & 38.3 & 60.3 & 41.4 & 62.4 \\
GFM-RAG+ \small (hidden\_dim=1024, 31M) \normalsize & 70.4 & 87.4 & 42.2 & 56.9 & 76.4 & 91.0 & 37.7 & 60.3 & 38.0 & 59.6 \\ 
GFM-RAG+ \small (hidden\_dim=2048, 122M) \normalsize & \textbf{73.1} & 88.1 & 44.0 & 57.9 & 77.5 & 91.8 & 39.0 & 60.5 & 35.3 & 57.0 \\ 
GFM-RAG+ \small (hidden\_dim=4096, 476M) \normalsize & \textbf{73.1} & 87.2 & 44.8 & 58.2 & \textbf{78.1} & 91.4 & 37.0 & 59.8 & 40.1 & 62.8 \\ \midrule
GLM-RAG \small (t5-small, 36M) \normalsize & 64.4 & 82.7 & 43.1 & 57.1 & 73.7 & 90.4 & 40.5 & 62.9 & 38.6 & 57.7 \\
GLM-RAG \small (t5-base, 110M) \normalsize & 68.8 & 85.9 & 44.8 & 61.1 & 76.1 & 91.9 & 43.6 & 66.6 & 48.2 & 66.6 \\ 
GLM-RAG \small (t5-large, 336M) \normalsize & 71.2 & 88.1 & \textbf{46.0} & \textbf{61.5} & 76.8 & 92.1 & \textbf{46.0} & \textbf{68.8} & \textbf{59.5} & \textbf{78.1} \\
\bottomrule
\end{tabular}
}
\end{table*}

\begin{table*}[b]
\centering
\caption{QA reasoning performance comparison (Exact Match and F1). \texttt{gpt-4o-mini} is used as the LLM. Best results are \textbf{bolded}, second best are \underline{underlined}, and third best are \textit{italized}. Baseline scores are adapted from \citet{greasoner}.}
\label{tab:qa_results}
\begin{tabular}{@{}lcccccc@{}}
\toprule
 & \multicolumn{2}{c}{HotpotQA} & \multicolumn{2}{c}{MuSiQue} & \multicolumn{2}{c}{2Wiki} \\ \cmidrule(l){2-7} 
Method & EM & F1 & EM & F1 & EM & F1 \\ \midrule
\multicolumn{7}{c}{\cellcolor[HTML]{C0C0C0}Non-structure Methods} \\
BM25 \citep{bm25} & 52.0 & 63.4 & 20.3 & 28.8 & 47.9 & 51.2 \\
ColBERTv2 \citep{colbertv2} & 43.4 & 57.7 & 15.5 & 26.4 & 33.4 & 43.3 \\
Qwen3-Emb (8B) \citep{qwen3} & 53.4 & 67.6 & 31.9 & 44.1 & 57.2 & 63.2 \\ \midrule
\multicolumn{7}{c}{\cellcolor[HTML]{C0C0C0}Graph-enhanced Methods} \\
RAPTOR \citep{raptor} & 50.6 & 64.7 & 27.7 & 39.2 & 39.7 & 48.4 \\
GraphRAG (MS) \citep{graphrag} & 51.4 & 67.6 & 27.0 & 42.0 & 34.7 & 61.0 \\
LightRAG \citep{lightrag} & 9.9 & 20.2 & 2.0 & 9.3 & 2.5 & 12.1 \\
KAG \citep{KAG} & \underline{59.5} & \underline{72.2} & \textit{33.8} & 46.0 & 67.3 & 75.1 \\
HippoRAG \citep{hipporag} & 46.3 & 60.0 & 24.0 & 35.9 & 59.4 & 67.3 \\
HippoRAG 2 \citep{hipporag2} & \textit{56.3} & \textit{71.1} & \underline{35.0} & \underline{49.3} & 60.5 & 69.7 \\
SubgraphRAG \citep{subgraphrag} & 44.5 & 57.0 & 25.1 & 35.7 & 62.7 & 69.0 \\
G-retriever \citep{gretriever} & 41.4 & 53.4 & 23.6 & 34.3 & 33.5 & 39.6 \\ 
\midrule
GFM-RAG \citep{main} & 56.2 & 69.5  & 30.2  & \textit{49.2} & \underline{69.8} & \underline{77.7} \\
GFM-RAG (reproduced) & 55.1 & 70.7 & 28.7 & 40.8 & 67.8 & 75.7 \\
GFM-RAG + & 55.3 & 70.6 & 31.6 & 43.3 & \textit{69.0} & \textit{77.0} \\
GFM-RAG* & 53.4 & 68.3 & 29.0 & 40.6 & 67.9 & 75.9 \\
GLM-RAG (ours) & 55.7 & 70.3 & 32.1 & 43.8 & 67.5 & 75.5 \\ \midrule
G-Reasoner \citep{greasoner} & \textbf{61.4} & \textbf{76.0} & \textbf{38.5} & \textbf{52.5} & \textbf{74.9} & \textbf{82.1} \\
\bottomrule
\end{tabular}
\end{table*}

\begin{table*}[b]
\centering
\caption{QA performance (F1 score) on out-of-domain datasets, answers generated with \texttt{gpt-4o-mini}.}
\label{tab:transferability_qa}
\resizebox{\textwidth}{!}{%
\begin{tabular}{lccccccc}
\toprule
Method & TechQA & PubMedQA & MS MARCO & HAGRID & ExpertQA & eManual & DelucionQA \\
\midrule
No Context & 19.1  & 20.8 & 29.9 & 21.5 & 16.7  & 19.6 & 19.1 \\
RAG Baseline & \textbf{25.4} & \textbf{21.7} & \textbf{36.7} & \textbf{27.1} & 19.4 & \underline{31.7} & 32.1 \\ \midrule
GFM-RAG* & 23.6 & 21.0 & 35.7 & 27.1 & 19.6 & 31.2 & \underline{33.6} \\
GFM-RAG & 23.5 & 21.1 & 35.3 & 26.9 & \underline{19.7} & 31.5 & \textbf{34.3} \\
GFM-RAG+ & 21.3 & 20.3 & 33.0 & 26.8 & 17.7 & 27.7 & 30.8 \\
GLM-RAG (ours) & \underline{25.2} & \underline{21.3} & \underline{35.9} & \underline{27.1} & \textbf{19.9} & \textbf{32.5} & 33.2 \\
\bottomrule
\end{tabular}
}
\end{table*}

\begin{figure*}[t]
    \centering
    \begin{tcolorbox}[title=Question Answering Prompt,
                      fontupper=\normalsize, breakable=false]
    \raggedright
\texttt{<system\_prompt>} \\
As an advanced reading comprehension assistant, your task is to analyze text passages and corresponding questions meticulously. Your response start after "Thought: ", where you will methodically break down the reasoning process, illustrating how you arrive at conclusions. Conclude with "Answer: " to present a concise, definitive response, devoid of additional elaborations.' \\
\texttt{<examples>} \\
input: |- \\
\textbf{Wikipedia Title}: Kurram Garhi \\
Kurram Garhi is a small village located near the city of Bannu, which is the part of Khyber Pakhtunkhwa province of Pakistan. Its population is approximately 35000. Barren hills are near this village. This village is on the border of Kurram Agency. Other nearby villages are Peppal, Surwangi and Amandi Kala. \\
\textbf{Wikipedia Title}: 2001–02 UEFA Champions League second group stage \\
Eight winners and eight runners- up from the first group stage were drawn into four groups of four teams, each containing two group winners and two runners- up. Teams from the same country or from the same first round group could not be drawn together. The top two teams in each group advanced to the quarter- finals. \\
\textbf{Wikipedia Title}: Satellite tournament \\
A satellite tournament is either a minor tournament or event on a competitive sporting tour or one of a group of such tournaments that form a series played in the same country or region. \\
\textbf{Wikipedia Title}: Trojkrsti \\
Trojkrsti is a village in Municipality of Prilep, Republic of Macedonia. \\
\textbf{Wikipedia Title}: Telephone numbers in Ascension Island \\
Country Code:+ 247\textless br\textgreater{} International Call Prefix: 00 Ascension Island does not share the same country code( +290) with the rest of St Helena. \\
\textbf{Question:} Are both Kurram Garhi and Trojkrsti located in the same country? \\
\textbf{Thought:} \\
response: |- \\
Kurram Garhi is located in the country of Pakistan. Trojkrsti is located in the country of Republic of Macedonia. Thus, they are not in the same country. \\
\textbf{Answer:} no. \\
\texttt{<doc\_prompt>} \\
"Wikipedia Title: \{title\} \\
\{content\}" \\
\texttt{<question>} \\
"Question: \{question\} \\
Thought: "
    \end{tcolorbox}
    \caption{The one-shot prompt template for 2Wiki. MuSiQue and HotpotQA have similar templates, adjusted to their examples. For the no-context prompt template, only the \texttt{<system\_prompt>} is given.}
    \label{fig:prompt}
\end{figure*}

\begin{table*}[t]
\centering
\newcommand{\ci}[2]{$#1_{\pm#2}$}
\caption{Retrieval recall (mean $\pm$ half the 95\% bootstrap CI width) on the question set shared by all systems within each dataset. Markers give a paired bootstrap test of GLM vs each system, corrected across all 88 comparisons with Holm-Bonferroni (FWER): $^{*}/^{**}/^{***}$ = GLM significantly better ($p<.05/.01/.001$), $^{\dagger}$ = significantly worse.}
\label{tab:error}
\resizebox{\textwidth}{!}{%
\begin{tabular}{l c c c c c c c c c c c}
\toprule
& \multicolumn{5}{c}{\textbf{Recall@2}} & & \multicolumn{5}{c}{\textbf{Recall@5}} \\
\cmidrule(lr){2-6}\cmidrule(lr){8-12}
Dataset & RAG & GFM-RAG & GNN & GNN$^{+}$ & GLM & & RAG & GFM-RAG & GNN & GNN$^{+}$ & GLM \\
\midrule
2WikiMultihopQA & \ci{.485}{.017}$^{***}$ & \ci{.790}{.016} & \ci{.790}{.017} & \textbf{\ci{.796}{.016}} & \ci{.785}{.016} &  & \ci{.594}{.017}$^{***}$ & \ci{.924}{.011} & \ci{.922}{.011} & \textbf{\ci{.930}{.011}} & \ci{.922}{.011} \\
DelucionQA & \textbf{\ci{.687}{.077}}$^{\dagger\dagger\dagger}$ & \ci{.423}{.107} & \ci{.473}{.098} & \ci{.460}{.093} & \ci{.500}{.097} &  & \textbf{\ci{.946}{.044}} & \ci{.717}{.100} & \ci{.747}{.087} & \ci{.770}{.088} & \ci{.813}{.083} \\
eManual & \textbf{\ci{.700}{.121}} & \ci{.514}{.148} & \ci{.581}{.145} & \ci{.457}{.143} & \ci{.600}{.140} &  & \textbf{\ci{.967}{.038}} & \ci{.676}{.140} & \ci{.686}{.133} & \ci{.776}{.119} & \ci{.848}{.098} \\
ExpertQA & \textbf{\ci{.638}{.052}}$^{\dagger\dagger\dagger}$ & \ci{.385}{.058}$^{*}$ & \ci{.403}{.056} & \ci{.462}{.055} & \ci{.445}{.056} &  & \textbf{\ci{.892}{.037}}$^{\dagger\dagger\dagger}$ & \ci{.581}{.064}$^{***}$ & \ci{.592}{.060}$^{***}$ & \ci{.670}{.057} & \ci{.677}{.056} \\
HAGRID & \textbf{\ci{.807}{.017}}$^{\dagger\dagger\dagger}$ & \ci{.629}{.023}$^{***}$ & \ci{.648}{.022}$^{***}$ & \ci{.681}{.020} & \ci{.696}{.021} &  & \textbf{\ci{.961}{.008}}$^{\dagger\dagger\dagger}$ & \ci{.815}{.019}$^{***}$ & \ci{.837}{.018}$^{***}$ & \ci{.882}{.015} & \ci{.889}{.015} \\
MS\,MARCO & \textbf{\ci{.374}{.022}}$^{\dagger\dagger\dagger}$ & \ci{.262}{.022}$^{*}$ & \ci{.260}{.022} & \ci{.284}{.021} & \ci{.295}{.021} &  & \textbf{\ci{.736}{.021}}$^{\dagger\dagger\dagger}$ & \ci{.565}{.028}$^{***}$ & \ci{.569}{.029}$^{***}$ & \ci{.604}{.026} & \ci{.630}{.027} \\
PubMedQA & \textbf{\ci{.603}{.016}}$^{\dagger\dagger\dagger}$ & \ci{.354}{.020}$^{***}$ & \ci{.364}{.019}$^{***}$ & \ci{.382}{.019}$^{***}$ & \ci{.437}{.019} &  & \textbf{\ci{.914}{.011}}$^{\dagger\dagger\dagger}$ & \ci{.582}{.021}$^{***}$ & \ci{.601}{.021}$^{***}$ & \ci{.610}{.021}$^{***}$ & \ci{.684}{.020} \\
TechQA & \textbf{\ci{.561}{.083}} & \ci{.290}{.073}$^{***}$ & \ci{.340}{.078} & \ci{.310}{.081} & \ci{.404}{.080} &  & \textbf{\ci{.715}{.073}} & \ci{.457}{.079}$^{***}$ & \ci{.482}{.080} & \ci{.480}{.080} & \ci{.571}{.078} \\
HotpotQA & \ci{.592}{.019}$^{***}$ & \ci{.728}{.020} & \ci{.733}{.021} & \textbf{\ci{.770}{.019}} & \ci{.745}{.019} &  & \ci{.730}{.019}$^{***}$ & \ci{.887}{.016} & \ci{.871}{.017} & \textbf{\ci{.906}{.014}} & \ci{.890}{.015} \\
MultiHop-RAG & \ci{.312}{.013}$^{***}$ & \ci{.341}{.019}$^{***}$ & \ci{.346}{.020}$^{***}$ & \ci{.390}{.019}$^{***}$ & \textbf{\ci{.600}{.019}} &  & \ci{.525}{.015}$^{***}$ & \ci{.537}{.021}$^{***}$ & \ci{.560}{.020}$^{***}$ & \ci{.639}{.020}$^{***}$ & \textbf{\ci{.775}{.017}} \\
MuSiQue & \ci{.332}{.016}$^{***}$ & \ci{.437}{.019}$^{***}$ & \ci{.439}{.019} & \ci{.456}{.019} & \textbf{\ci{.463}{.018}} &  & \ci{.463}{.017}$^{***}$ & \ci{.577}{.020}$^{***}$ & \ci{.578}{.020}$^{***}$ & \ci{.608}{.020} & \textbf{\ci{.620}{.019}} \\
\bottomrule
\end{tabular}%
}
\end{table*}

\end{document}